\documentclass[journal]{IEEEtran}
\usepackage{cite}
\usepackage{amsmath,amssymb,amsfonts}
\usepackage{hyperref}
\usepackage{graphicx}
\usepackage{textcomp} 
\usepackage{multicol}
\usepackage{multirow}
\usepackage{subfigure}
\usepackage[dvipsnames]{xcolor}
\usepackage{pifont}
\usepackage{lettrine}
\usepackage{booktabs}
\usepackage{amsmath,amssymb,amsfonts}
\usepackage{hyperref} 
\usepackage{soul}
\usepackage{graphicx}
\usepackage{tikz}
\usepackage{adjustbox}
\usepackage{svg}
\usepackage{array}       

\newcommand{\pmark}{\textcolor{orange}{$\approx$}} 
\newcommand{\cmark}{\textcolor{ForestGreen}{\ding{51}}}
\newcommand{\xmark}{\textcolor{red}{\ding{55}}}
\definecolor{LightRed}{rgb}{1,0.92,0.92}
\definecolor{LightOrange}{rgb}{1,0.95,0.88}
\definecolor{LightYellow}{rgb}{1.0,1.0,0.84}
\definecolor{LightGreen}{rgb}{0.9,1.0,0.88}
\definecolor{LightCyan}{rgb}{0.9,1,1}
\definecolor{LightBlue}{rgb}{0.9,0.94,1}
\definecolor{LightIndigo}{rgb}{0.92,0.9,1}
\definecolor{LightMagenta}{rgb}{0.96,0.86,1}
\definecolor{DirtyWhite}{rgb}{0.96,0.96,0.96}

\providecommand{\cmark}{\textcolor{black!75}{\ding{51}}}            
\providecommand{\xmark}{\textcolor{black!40}{\ding{55}}}            
\providecommand{\pmark}{\textcolor{black!60}{\ding{108}}}           
\newcommand{\gmark}{\textcolor{black!75}{\ding{119}}}               

\newcommand{\osi}{\textsc{osi}}      
\newcommand{\lcl}{\textsc{ccl}}      
\newcommand{\ncl}{\textsc{nc}}       
\newcommand{\ndl}{\textsc{nc-nd}}    
\newcommand{\bsp}{\textsc{bsp}}      
\newcommand{\nol}{\textsc{none}}     
\newcommand{\nal}{\textsc{n/a}}      

\title{Open Foundation Models in Healthcare:\\ Challenges, Paradoxes, and Opportunities with GenAI Driven Personalized Prescription}

\author{Mahdi Alkaeed$^{1}$, Sofiat Abioye$^{2}$, Adnan Qayyum$^{3}$, Yosra Magdi Mekki$^{4}$, Ilhem Berrou$^{5}$, Mohamed Abdallah$^{3}$,\\ Ala Al-Fuqaha$^{3}$, Muhammad Bilal$^{2}$, and Junaid Qadir$^{1}$ \\
$^1$Department of Computer Science and Engineering, College of Engineering, Qatar University, Doha, Qatar \\
$^2$Birmingham City University, Birmingham, United Kingdom \\
$^3$College of Science and Engineering, Hamad Bin Khalifa University (HBKU), Doha, Qatar \\
$^4$College of Medicine, Qatar University, Doha, Qatar \\
$^5$University of the West of England (UWE), Bristol, United Kingdom\\
}

\begin{document}

\maketitle

\begin{abstract}
    
In response to the success of proprietary Large Language Models (LLMs), there is a growing interest in developing open, non-proprietary LLMs and AI foundation models (AIFMs) for transparent use in academic, scientific, and non-commercial applications. Open models have substantial potential to improve healthcare applications. In this paper, we provide a comprehensive survey and a taxonomy of state-of-the-art open LLMs and AIFMs for clinical tasks. In addition, we empirically evaluate open-weight LLMs on a personalized-prescription task and compare them with a proprietary model with and without Retrieval-Augmented Generation (RAG). Across 100 synthetic patient profiles, the open-weight models matched or exceeded the proprietary baseline on overall suitability accuracy without retrieval augmentation (0.810--0.850 versus 0.780), although the confidence intervals overlap; with retrieval augmentation, every model improved, and the ranking closed to within 0.04. Gains from RAG were field-dependent rather than uniform. To highlight the clinical relevance of LLM-driven personalized prescriptions, we also conduct a subjective evaluation in which two raters, blinded to model and condition, scored responses independently of each other. Their ratings did not follow the accuracy gains: retrieval raised overall suitability accuracy for every model while lowering the two-rater means for two of the four on all five dimensions. Finally, we outline the ethical implications and potential risks associated with the misuse of advanced LLMs and AIFMs, highlighting the importance of responsible and transparent deployment in healthcare contexts. \textit{Data and code used in the paper will be released upon publication.}
\end{abstract}
\begin{IEEEkeywords}
AI Foundation Models (AIFMs), Large Language Models (LLMs), Retrieval-Augmented Generation (RAG), Personalized Prescriptions, LLMs for Healthcare.
\end{IEEEkeywords}

\section{Introduction}
Recent years have witnessed an upsurge in the development of various \textbf{A}rtificial \textbf{I}ntelligence \textbf{F}oundation \textbf{M}odels (AIFMs) for tasks such as Natural Language Processing (NLP), computer vision, and healthcare. The key driving force behind these models is the advancements in Generative Artificial Intelligence (GenAI), which have demonstrated significant potential across different domains, including healthcare, offering opportunities to improve patient care, enable personalized treatment strategies, support professional training, and accelerate medical research \cite{bhuyan2025generative}. This trend is primarily driven by the availability of extensive datasets and advancements in Deep Learning (DL), such as transformer networks. Notable milestones include the Transformer (213 million parameters in its large configuration) \cite{vaswani2017attention}, GPT-2 (1.5 billion parameters) \cite{radford2019language}, GPT-3 (175 billion parameters) \cite{brown2020language}, and the Switch Transformer (1.6 trillion parameters) \cite{fedus2022switch}, and the GPT-5 series (undisclosed parameters). Specifically, scaling of model sizes has resulted in significant improvements in the performance of these AIFMs across classical NLP benchmarks such as GLUE \cite{wang2018glue}, SuperGLUE \cite{wang2019superglue}, and Winograd \cite{sakaguchi2021winogrande}. Furthermore, AIFMs have showcased their efficacy as few-shot learners \cite{brown2020language}, fueling their widespread adoption as foundational pre-trained models \cite{chen2020recall,chua2021fine,ju2022robust,vulic2021lexfit,zhang2020revisiting}.

\begin{table*}
\centering
\footnotesize
\caption{Comparison of related works on foundation models in healthcare with our paper. 
Legend: OS $\rightarrow$ Open Source; FMs $\rightarrow$ Foundation Models; Tax. $\rightarrow$ Taxonomy; 
Emp. CS $\rightarrow$ Empirical Case Study; 
OS vs. CS $\rightarrow$ Open-Source vs. Closed-Source benchmarking; 
Human $\rightarrow$ human or expert evaluation of model outputs reported; \cmark $\rightarrow$ covered; \pmark $\rightarrow$ partially covered; \xmark $\rightarrow$ not covered.}

\label{tab:related}
\scalebox{0.75}{%
\begin{tabular}{p{2.3cm} p{2.8cm} p{2cm} p{8cm} c c c c  c}
\toprule
\textbf{Ref.} & \textbf{Type} & \textbf{Domain} & \textbf{Primary Focus} 
& \textbf{OS} & \textbf{OS FM Tax.} & \textbf{Emp. CS} & \textbf{OS vs. CS}  & \textbf{Human} \\
\midrule
He et al.~\cite{he2024foundation} &
Methodological review & Healthcare & Provides a comprehensive review of healthcare FMs, including multi-modal FMs. & \xmark & \pmark & \xmark & \xmark & \xmark \\ \midrule

Khan et al.~\cite{khan2025comprehensive} & Methodological review & Medicine & Provides a taxonomy of FMs in medicine and discusses their challenges & \xmark & \pmark & \xmark & \xmark & \xmark \\ \midrule

Mahesh et al.~\cite{mahesh2024advancing} & Review & Healthcare & Explores the adoption of FMs in healthcare along with challenges and ethical considerations. & \xmark & \xmark & \xmark & \xmark & \xmark \\ \midrule

Li et al.~\cite{li2025open} & Review & Biomedical Healthcare & Surveys FMs with federated learning for biomedical applications. & \pmark & \pmark & \xmark & \xmark  & \xmark \\ \midrule

Zhang et al.~\cite{zhang2026data} & Survey & Computational healthcare (CH) & Data-centric FMs in CH applications and associated challenges & \xmark & \pmark & \xmark & \xmark & \xmark \\ \midrule

Bilal et al.~\cite{bilal2025foundation} & Survey & Pathology & Overviews applications of FMs in computational pathology & \xmark & \xmark & \xmark  & \xmark & \xmark \\ \midrule

Huang et al.~\cite{huang2025systematic} & Systematic review & Medical imaging & Reviews multimodal FMs with implementation perspectives & \xmark & \xmark & \xmark & \xmark & \xmark \\ \midrule

Rajendran et al.~\cite{rajendran2026foundation} & Systematic review & Medical imaging & Unified synthesis and meta-analysis of FMs in medical image analysis & \xmark & \pmark & \xmark & \xmark & \xmark \\ \midrule

Chia et al.~\cite{chia2024foundation} & Review & Ophthalmology & Applications of foundation models in ophthalmology & \xmark & \xmark & \xmark & \xmark & \xmark \\ \midrule

Phuntsho et al.~\cite{phuntsho2025adaptation} & Survey & Medical Imaging & Discusses adaptation strategies for adapting FMs for medical imaging. & \xmark & \xmark & \pmark & \xmark & \xmark \\ \midrule

Templin et al.~\cite{templin2025framework} & Framework/review & Healthcare LLMs & Five-step audit framework for accuracy and bias evaluation of clinical LLMs. & \xmark & \xmark & \pmark & \xmark & \pmark \\ \midrule

\textbf{This paper} & Survey + empirical & Healthcare & Explicitly focuses on OS FMs in healthcare, provides a taxonomy, and an empirical case study on personalized prescription to benchmark open and closed LLMs under blinded two-rater review. & \cmark & \cmark & \cmark & \cmark & \cmark \\ \bottomrule
\end{tabular}}
\end{table*}

Following the success of ChatGPT driven by proprietary Large Language Models (LLMs) like GPT-4 from OpenAI, there is increasing interest in open and nonproprietary LLMs. These models are positioned for broader community use in academic, scientific, and noncommercial endeavors. Although some open-weight AI models have emerged, the landscape remains in flux, and model selection increasingly depends on performance, licensing, and integration capabilities. Although these open models often do not match the refined functionality of proprietary counterparts like GPT-4, they promise to revitalize healthcare applications when tailored to specific tasks. For example, open AIFMs can be repurposed to analyze vast medical literature, clinical guidelines, and patient data, helping healthcare professionals make more accurate diagnoses by considering a broader range of information and patterns than human capabilities alone, ultimately improving healthcare outcomes. Nevertheless, various lingering doubts and ethical considerations arise, particularly regarding the potential misuse of such powerful models for malicious purposes.  

This paper emphasizes the latest advancements and practical applications of open LLMs and AIFMs in healthcare. We evaluate the impact of open AI on healthcare and present the opportunities, capabilities, and challenges presented by current open AIFMs in this field. A concrete definition of open AI models is currently unavailable in the literature. Seger et al. \cite{seger2023open} defined open AI models as models for which the underlying training data, source code, and model weights are publicly available. On the other hand, Kapoor et al. \cite{kapoor2024societal} followed a reductive approach, defining open AI models as those with widely available weights. In addition, the security, robustness, and privacy of these models remain unaddressed, as they are not as polished as proprietary closed-source models and lag behind the current capabilities of proprietary closed-source LLMs and AIFMs \cite{thieme2023foundation,seger2023open}.

\textit{Contributions of the Paper:} This article surveys open AIFMs and LLMs for healthcare, with particular emphasis on model openness, accessibility, and practical use. Whereas prior surveys examine healthcare foundation models more broadly (Table~\ref{tab:related}), this work combines a focused survey with a technical case study. Specifically, we develop a taxonomy of open-weight models and conduct a paired empirical evaluation of open-weight and proprietary LLMs.  Our scope is also distinct from complementary research on bias auditing in clinical LLMs \cite{templin2025framework}. The major contributions of this paper are as follows:

\begin{enumerate}
    \item We provide a comprehensive survey of open-weight LLMs and AIFMs in healthcare, synthesizing recent advances, application areas, and emerging research directions. 
    \item We develop a taxonomy of open healthcare AIFMs organised by application area and by degree of release openness.
    \item We present a technical case study comparing open-weight and proprietary LLMs on clinical prescription and adverse-risk assessment tasks. The evaluation includes two general-purpose open-weight LLMs (Llama 4 Scout and Mistral Small 3.2 24B) and one medical-domain open-weight LLM (MedGemma 27B), benchmarked against the proprietary GPT-5.6 Luna model. We also investigate how integrating Retrieval-Augmented Generation (RAG) affects the performance of these LLMs in this critical clinical task.
\end{enumerate}

Before moving further, it is worth defining the term ``AIFMs'' and how they are different from LLMs. AIFMs are large, general-purpose models typically trained with self-supervision and heterogeneous, large-scale data to learn transferable representations and generative capabilities across modalities. AIFMs include vision backbones, diffusion generators, and vision-language encoders or decoders. They support adaptation through prompting, instruction tuning, or task-specific fine-tuning. LLMs are a subset of foundation models specialized for natural language; they are trained primarily on text and are optimized for tasks such as question answering, dialogue, summarization, and information extraction. In short, LLMs are text-centric foundation models, whereas AIFMs span broader modality coverage (e.g., vision, audio, and multimodal settings) and enable cross-modal representation learning and generation. In this paper, we use ``AIFMs'' to refer to multimodal foundation models and ``LLMs'' to refer to text-only models.

\begin{figure*}
\centering
\includegraphics[width=\linewidth]{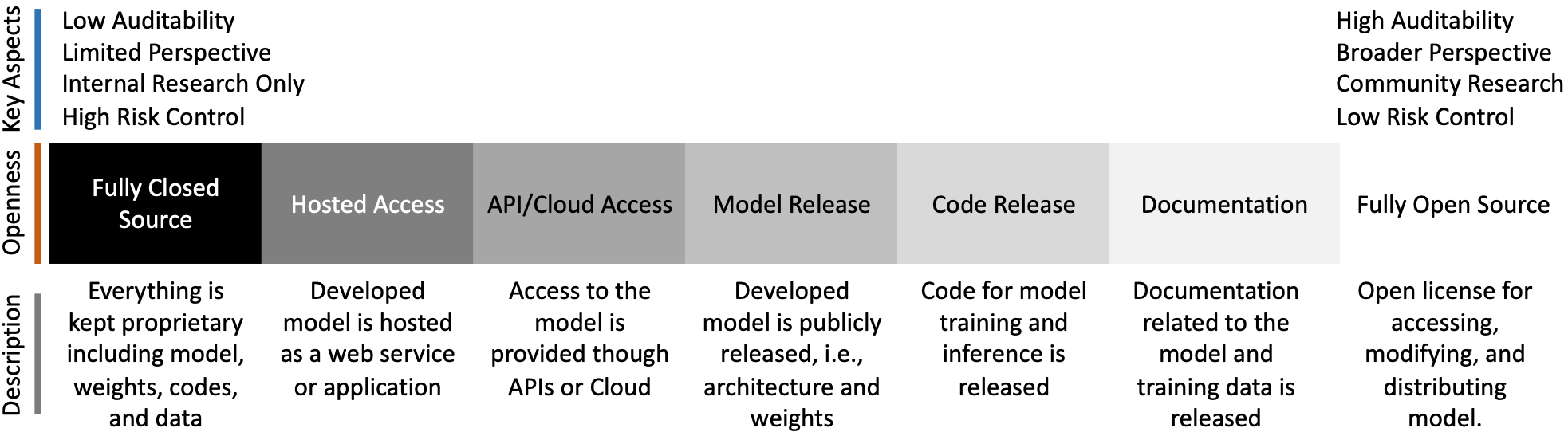}
\caption{\textbf{Levels of Open-Source Access for LLMs and AIFMs.} The figure shows the spectrum from Fully Closed Source to Fully Open Source, detailing key aspects and descriptions for each level. (Adapted from \cite{solaiman2023gradient}).}
\label{fig:openness}
\end{figure*}

\textit{Organization of the Paper:}
The remainder of the paper is structured as follows: Section \ref{sec:back} provides an overview of the relevant background on open-source LLMs and AIFMs, along with their associated challenges. Section \ref{sec:sota_review} provides an overview of the existing state-of-the-art literature on open LLMs and AIFMs in the healthcare domain. Section \ref{sec:case_study} presents the methodology and results of our case study on personalized prescriptions. Finally, we conclude the paper in Section \ref{sec:concs}.

\section{Background}
\label{sec:back}
\subsection{LLMs and AI Foundation Models: A Brief Overview}
LLMs represent recent advances in large, general models characterized by their immense size and capacity to perform a wide range of NLP tasks, including understanding and generating human-like text \cite{vaswani2017attention}. LLMs are typically pre-trained on large datasets containing vast amounts of text from the Internet. During pre-training, the model learns to predict the next word in a sentence or fill in missing parts of text. This unsupervised pre-training allows the model to capture contextual information and linguistic structures. After pre-training, LLMs can be fine-tuned on specific tasks with smaller datasets. Fine-tuning adapts the model to perform tasks such as text classification, language translation, summarization, etc. These models can generate coherent, contextually relevant text in response to a prompt. LLMs have been used for various downstream tasks, including creative writing, code generation, question answering, and other natural language understanding tasks \cite{he2025survey}. The success of LLMs across various NLP tasks has attracted substantial attention from the research community and led to the development of AIFMs, which extend the same paradigm beyond text to vision and multimodal settings, providing transferable representations and generative abilities that can be specialized for diverse applications. In healthcare, both LLMs and AIFMs support capabilities such as clinical note summarization, longitudinal patient data analysis, decision support for diagnostics, and personalized treatment recommendations, while offering measurable potential to improve efficiency and quality of care \cite{bommasani2021opportunities}.

\subsection{Open Source LLMs and AI Foundation Models}
What it means for a model to be open, however, is far less settled than the enthusiasm for openness suggests. The concept of Open AI, or open-source AI, remains nebulous, often serving as an aspiration or marketing tactic rather than a concrete definition \cite{seger2023open}. These terms reflect the principles of free and open-source software (FOSS), emphasizing the freedom to use, study, modify, and share software. Open LLMs and AIFMs extend this philosophy to AI, highlighting the importance of transparency, accessibility, and collaborative enhancement of these models. The decision between open-source and closed-source models involves balancing the freedom for customization against proprietary features and commercial support offered by closed-source models. Open LLMs and AIFMs are distributed under various licenses \cite{carugati2023competition}, from the permissive MIT or Apache 2.0 (which allows considerable user freedom) to the more restrictive GNU General Public License (GPL) \cite{gnu_gpl3}, which requires distributed modified versions to be licensed under the GPL. These licenses facilitate commercial use, modification, and distribution, while also imposing restrictions on copyright and patent rights. In the ML research and development community, doubts remain about the merits and drawbacks of making LLMs and AIFMs open-source or retaining them as proprietary models \cite{alizadeh2025open,seger2023open}.

\subsubsection{Requirements for Open LLMs and AIFMs}
The term open-source stems from the traditional Open-Source Software (OSS) context, defined in 1998 \cite{osi1998definition}. OSS requires software to be publicly available for use, viewing, modification, and distribution of source code under an open-source license \cite{seger2023open, dibona1999open}. However, within the context of LLMs and AIFMs, the term ``open AI'' has become ambiguous. The degree of openness in existing open AI tools has been a subject of considerable debate. For example, LLaMA has encountered criticism regarding the extent of their openness \cite{liesenfeld2024rethinking}. For instance, Meta recently announced an open-source LLaMA-2 model; however, it has not been made available under a traditional open-source license. Instead, Meta has defined its terms and conditions governing the use of this model across various IoT applications, especially in healthcare. In Figure \ref{fig:openness}, we present a gradient depicting the level of open-source access to LLMs and AIFMs, followed by a brief discussion of the pros and cons of open-sourcing these models.

The requirements for open AI are certainly different, and various definitions are found in the relevant literature. For example, Seger et al. \cite{seger2023open} outlined the critical components of open-source AI models, encompassing everything from training data to the model itself. In contrast, Kapoor et al. \cite{kapoor2024societal} argued that open access to model weights alone suffices to classify a model as open AI. However, several components are involved in developing AI models, including model architecture, training data, hyperparameters, model weights, and documentation. Each of these components contributes to the overall transparency and functionality of the models. In common practice, developers often have the flexibility to choose which components of the open AI models they make public, which allows them to tailor their open-source approach based on the specific considerations of their project, taking into account factors such as proprietary concerns, collaboration preferences, and the nature of the AI application being developed. However, in the context of LLMs and AIFMs, the crucial open-source requirement is to provide open access to their weights. Therefore, releasing only one component of a model, such as the source code, does not constitute a genuinely open-source model. This is the operational rule used throughout the survey and taxonomy: weight availability is necessary; code-only or documentation-only releases are excluded; fully open releases (weights, code, data, and documentation) lie at the open end of Figure~\ref{fig:openness}.

In October 2024, the Open Source Initiative published version 1.0 of the Open Source AI Definition (OSAID) \cite{osi2024osaid}, the first formal standard for openness in AI systems. The OSAID requires three components: sufficiently detailed information about the training data for a skilled person to build a substantially equivalent system, the complete training and inference source code, and model parameters available under OSI-approved terms. Section~\ref{sec:protocol} describes how we apply the OSAID to score each model identified in our survey. There are various prospects for using open AI models in the healthcare domain, including enhancing medical research, improving diagnostic tools, and facilitating personalized treatment plans. For instance, open-source models could analyze large datasets of medical literature to identify new treatment protocols or assist clinicians by providing evidence-based recommendations tailored to individual patient needs.

\subsubsection{Pros and Cons of Using Open AIFMs}
Seger et al. \cite{seger2023open} described the key advantages of using open LLMs and AIFMs across three dimensions including: (1) \textit{External Evaluation:} Open-sourcing LLMs and AIFMs provides an opportunity for the wider AI community to perform external audits and extensive evaluations on the AI system to identify its vulnerabilities to enhance the overall safety and robustness of models; (2) \textit{Acceleration in AI Progress:} Open-sourcing these models can drive significant advancements through collective and widespread collaboration among AI developers and researchers; and (3) \textit{Distributed Control:} Open-sourcing LLMs and AIFMs can decentralize AI development, allowing smaller organizations to contribute. This approach democratizes AI but poses potential dangers, such as risks regarding the misuse of these powerful models \cite{seger2023open}.

In a recent study, Kapoor et al. \cite{kapoor2024societal} provided a comprehensive evaluation of the societal impact of open AIFMs. They identified five distinctive properties of these models: broader access, greater customizability, the potential for local inference, the inability to rescind model access once released, and weaker monitoring of the models' downstream use. The authors highlighted five significant benefits of open AIFMs: distributing decision-making power, reducing market concentration, increasing innovation, accelerating science, and enabling transparency. They emphasized that broader access and greater customizability facilitate innovation, enabling these models to be aggressively tailored to support diverse applications and advance the state of the art across languages. Furthermore, Kapoor et al. \cite{kapoor2024societal} introduced the concept of ``marginal risk'', which refers to the extent to which these models increase societal risk through intentional misuse, compared with existing closed AIFMs or other technologies such as web search. They also highlighted the challenge of enforcing restrictions on downstream usage in open-weight models, noting that the open release of model weights is irreversible.

\subsection{Challenges Hindering Open AI in Healthcare}
Conducting a comprehensive evaluation of LLMs across diverse healthcare roles, especially in high-stakes clinical scenarios, remains challenging. Current benchmarks are largely text-based, English-centric, and predominantly focused on medications, limiting their capacity to assess broader healthcare knowledge and multimodal reasoning \cite{liu2025kokushimd}. In \cite{shool2025systematic}, the authors presented a systematic review of the evaluation parameters and methodologies used for LLMs in clinical medicine, highlighting their capabilities, limitations, and application trends. They identified 761 studies that met the inclusion criteria, revealing a growing interest in leveraging LLM tools in clinical settings. General-domain LLMs, particularly GPT-4, dominated evaluations (93.55\%), while medical-domain LLMs accounted for only 6.45\%. Accuracy emerged as the parameter most frequently assessed (21.78\%). Despite these advancements, several challenges hinder the effective deployment of Open AI in healthcare, including the limited availability of specialized medical LLMs, insufficient comprehensive evaluation frameworks, and difficulties in addressing clinical safety, privacy concerns, data security risks, and the sensitive nature of patient information, which further complicate the reliable assessment and deployment of LLMs in clinical settings.

\subsubsection{Data and Model Privacy}
In the literature, it has been argued that ensuring the privacy of patient data is a top priority \cite{khalid2023privacy}. AIFMs are typically trained on large amounts of medical records, which may include sensitive, confidential, or personally identifiable information. Therefore, during training, the model can inadvertently memorize and reproduce this information when generating text, leading to data leakage \cite{inan2021training}. This can be a severe breach of privacy if the generated content is shared publicly or accessed by unauthorized/malicious entities. These extracted instances include publicly accessible personally identifiable information, such as names, phone numbers, and email addresses, as well as medical records. Larger models tend to exhibit greater vulnerability compared to smaller ones \cite{carlini2021extracting,inan2021training}. Proper encryption, access controls, and compliance with healthcare regulations are essential to protect patient information \cite{torkzadehmahani2022privacy}. Training healthcare AIFMs necessitates access to large quantities of high-dimensional data, often derived from unfiltered user-generated content. This approach introduces substantial privacy risks \cite{carlini2021extracting, inan2021training, zou2020privacy, pan2020privacy, goldstein2023generative, bender2021dangers}.

\subsubsection{Model Security}
Researchers have demonstrated that AI models used in medical imaging can be vulnerable to adversarial attacks. For instance, the behavior of an AI diagnostic model when presented with adversarial images generated by Generative Adversarial Networks (GANs) could lead it to misdiagnose conditions such as cancer. This type of attack could potentially lead to incorrect treatment plans, posing severe risks to patient health \cite{zhou2021machine}. Various adversarial ML attacks can be realized on DL models that can be broadly categorized into two classes \cite{qayyum2020secure}: (1) targeted attacks, where the objective is to force the model to classify a given input into the target class; (2) untargeted attacks, where the objective is to increase overall misclassification rate. The magnitude of these attacks has already been demonstrated across various ML-empowered medical systems, such as medical image classification \cite{finlayson2018adversarial,finlayson2019adversarial,paschali2018generalizability}, abnormal heartbeat detection (i.e., ECG classification) \cite{han2020deep}, etc. Although open-sourcing AI models will expedite research in critical directions such as evaluating their security aspects, a cautious approach should be considered while openly releasing models trained using patients' sensitive data.

\subsubsection{Ethical Issues}
Healthcare applications are essentially human-centric and require careful consideration of ethical aspects throughout the development of AI-based medical systems. Therefore, it is imperative to understand the sociological needs of the target users before beginning data collection to develop the AI model. In addition, the ethical dimensions of AI applications in healthcare involve addressing biases, ensuring fairness, and contemplating the broader implications of AI-generated decisions on patient care \cite{wang2023ethical}. John et al. \cite{john2024re} argued that ethical debates need to be localized within the complex interplay of technical, legal, and organizational entities from which ML moral issues arise.

\subsubsection{Regulatory Issues}
The true transformative potential of AI in healthcare can only be realized when these systems are effectively integrated into clinical workflows. The healthcare sector operates under stringent regulatory frameworks, such as the Health Insurance Portability and Accountability Act (HIPAA) \cite{oakley2023hipaa} and the General Data Protection Regulation (GDPR) \cite{carmi2023european}, which impose strict requirements on patient privacy, data handling, and model transparency. Integrating AI systems within these frameworks requires careful alignment between technological capabilities and legal obligations. Beyond regulatory challenges, the generation of clinical prescriptions itself embodies considerable complexity, as it involves synthesizing heterogeneous patient data, including comorbidities, laboratory results, and ongoing treatments, while taking into account drug-drug and drug-disease interactions, dosage adjustments, and adherence to evidence-based clinical guidelines. These layers of reasoning underscore that the safe deployment of AI-driven decision support is far from straightforward. The literature further highlights that current regulatory guidelines remain insufficient to oversee AI-based medical systems, as these models are adaptive and continually evolve with exposure to new patient data \cite{qayyum2020secure}. Consequently, there is a pressing need for dynamic governance frameworks and robust validation pipelines that ensure AI solutions remain safe, explainable, and compliant as they are integrated into real-world healthcare settings.

\subsection{Data Governance and Model Auditing}
Beyond privacy and security safeguards, the sustainable deployment of open healthcare AI models depends on rigorous data governance and transparent auditing practices. This involves maintaining detailed documentation such as model cards and data statements to disclose dataset provenance, curation methodology, and known limitations. For privacy-sensitive applications, federated learning and synthetic data generation have emerged as key strategies for training models without centralizing identifiable patient data. Federated approaches, in particular, ensure that sensitive health records remain within institutional boundaries while contributing to collaborative model improvement. Furthermore, model auditing frameworks should be institutionalized to assess fairness, performance drift, and reproducibility across demographic and clinical subgroups, ensuring that open models remain trustworthy and aligned with evolving healthcare standards~\cite{gebru2021datasheets,mitchell2019model}.

\section{State-of-the-Art in Open Healthcare LLMs and AIFMs}
\label{sec:sota_review}
\subsection{Applications of LLMs and AIFMs in Healthcare}
\begin{figure*}
\centering
\includegraphics[width=0.95\linewidth]{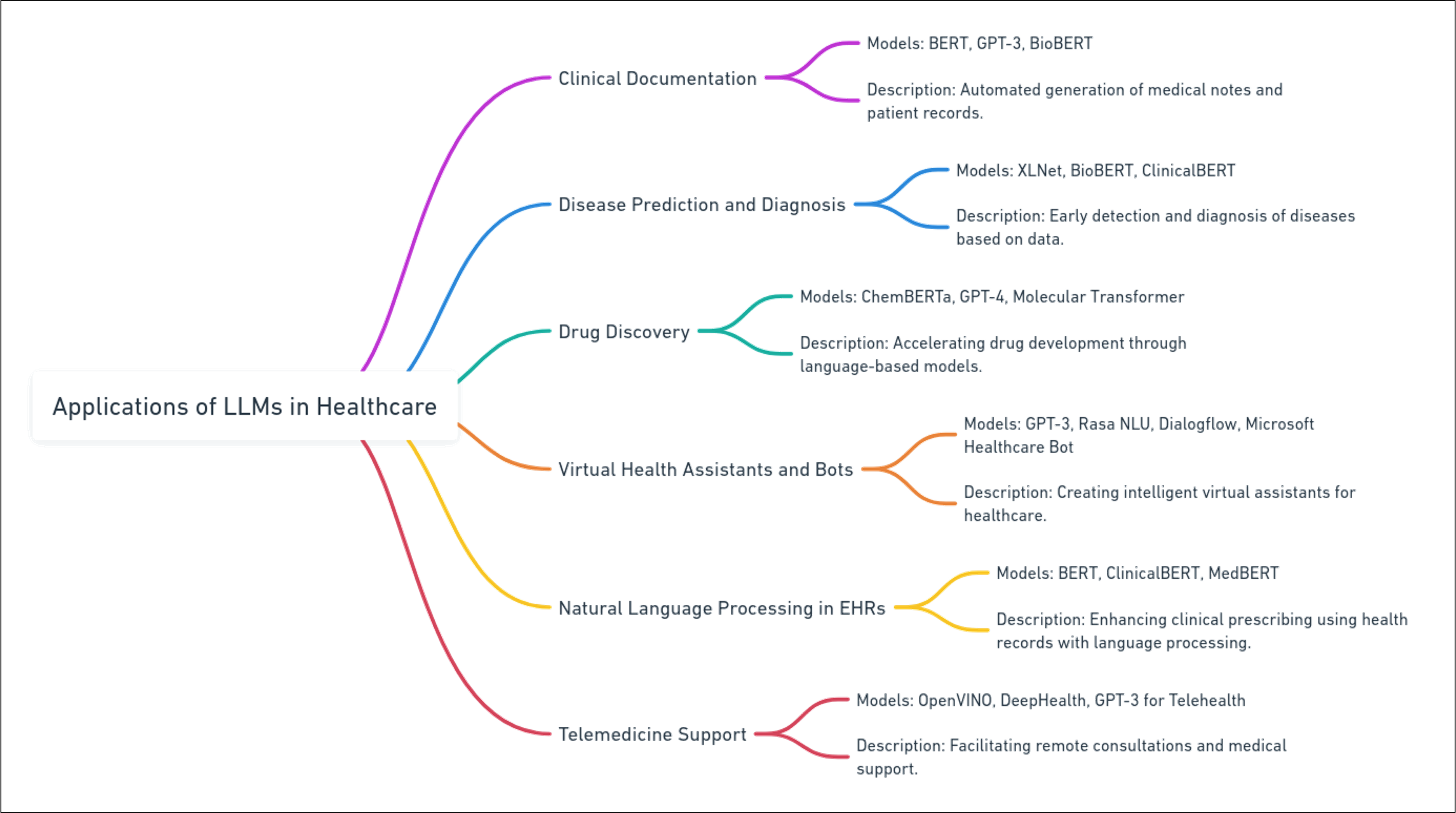}
\caption{\textbf{Applications of LLMs in Healthcare.} This figure shows six key areas: Clinical Documentation, Disease Prediction and Diagnosis, Drug Discovery, Virtual Health Assistants, NLP in EHRs, and Telemedicine Support.}
\label{fig:ApplicationsOfLLMs}
\end{figure*}

LLMs hold considerable promise for advancing healthcare across clinical, operational, and research domains. Their emergence has facilitated the exploration of medical applications by enabling sophisticated language understanding, reasoning, and communication in healthcare contexts \cite{vrdoljak2025review}. Notable examples, such as ChatGPT and DeepSeek-R1, demonstrate advanced natural language processing and reasoning capabilities, positioning them to enhance both the quality and efficiency of healthcare services \cite{wang2025novel}. This development holds the promise of more immersive human-AI interaction and collaboration. Notably, these models have demonstrated remarkable proficiency on multiple-choice research benchmarks \cite{wei2022chain,singhal2023large}. Moreover, multimodal foundational models can learn visual features from textual descriptions and are adept at tasks such as medical imaging analysis, disease detection in radiology, and pathology slide interpretation. Leveraging state-of-the-art architectures like convolutional neural networks (CNNs) and transformer-based models such as CLIP \cite{radford2021learning}, language vision models can enhance diagnostic accuracy, streamline medical imaging workflows, and contribute to the development of advanced healthcare technologies such as VisionFM \cite{qiu2024development}, RETFound \cite{zhou2023foundation}, and LVM-Med \cite{mh2024lvm}. Figure \ref{fig:ApplicationsOfLLMs} illustrates a spectrum of LLMs' applications in healthcare.

Moreover, multimodal LLMs can analyze and process data from various resources and modalities, including text (language), images (vision), and audio. The idea is to create models that can understand and generate insights by jointly considering information from different types of data. LLMs are reshaping a broad range of healthcare tasks. LLMs and AIFMs offer a versatile set of applications that significantly impact various facets of healthcare. From automating clinical documentation to early disease prediction, for instance, LLMs such as BERT \cite{devlin2019bert}, BioBERT \cite{lee2020biobert}, and ChatGPT \cite{vaishya2023chatgpt} have demonstrated the capability to process complex medical information efficiently. In addition, these models can be leveraged for drug discovery, personalized medicine, and the automation of literature reviews, illustrating their potential to accelerate research and enhance patient care. The integration of LLMs and multimodal AIFMs into healthcare practices underscores their capacity to address diverse requests, making them valuable assets for advancing information processing and decision-making \cite{clusmann2023future}.

\subsubsection{Applications in EHR Data Processing}
LLMs offer a diverse range of applications, including automating clinical documentation to enhance accuracy and extracting structured data from unstructured EHR data \cite{wornow2023shaky}. It also facilitates streamlining coding and billing processes, providing real-time insights for clinical decision support, improving information search through NLP, facilitating population health management through trend analysis, enabling voice-to-text transcription for efficient record updates, enhancing patient engagement with personalized summaries, ensuring data quality through automated quality assurance, and supporting healthcare research with advanced analytics. The integration of LLMs into EHRs is therefore central to optimizing healthcare workflows, enhancing decision-making, and ultimately improving patient outcomes \cite{wang2023anypredict,guo2023ehr}. These uses share structured extraction and decision support over EHR fields rather than unconstrained generation.

\subsubsection{Applications in Clinical NLP}
Various NLP techniques are used to extract structured information from EHRs, converting free-text data into a format that can be easily analyzed for research, decision support, and other healthcare applications. Clinical NLP applications are diverse and impactful in healthcare \cite{singhal2023large}. They include improving clinical documentation accuracy, extracting structured information from EHRs, aiding clinical decision support, extracting phenotypic information for research, detecting adverse drug events, expediting clinical trial recruitment, understanding temporal relationships in patient records, de-identifying data for privacy compliance, and facilitating voice-to-text transcription in clinical settings \cite{wornow2024zero}. Powerful LLMs like GPT-4 demonstrate significant potential to process text data within the medical domain through zero-shot in-context learning \cite{liu2023deid,waisberg2023gpt}. Clinical notes exhibit significant differences from biomedical literature, leading to subpar performance of BERT models pre-trained on clinical notes, similar to the standard BERT model \cite{gu2021domain}. The unifying task is converting unstructured notes into analyzable structure; domain shift from biomedical literature remains a bottleneck.

\subsubsection{Applications in Prognosis, Diagnosis, and Medicine}
LLMs play a crucial role in healthcare across different applications, including prognosis, diagnosis, and medicine \cite{sun2024pathasst,balas2023conversational}. In the prognosis task, they aid in predicting outcomes and modeling disease progression. Similarly, LLMs can provide automated diagnostic support \cite{levine2023diagnostic}, analyze medical images for abnormalities, and assist in identifying rare diseases. In medicine, LLMs can be leveraged for drug discovery, enable personalized treatment plans, and help monitor and improve medication adherence \cite{xie2023faithful}. Prognosis, diagnosis, and treatment planning all depend on combining textual evidence with patient-specific context.

\subsubsection{Applications in Medical Imaging}
Vision-centric foundation models are reshaping core tasks in medical imaging, including classification, detection, segmentation, and vision--language grounding across various modalities such as X-ray, CT, MRI, ultrasound, and pathology. For instance, one of the famous vision models for image segmentation, namely Segment Anything Model (SAM) has been adapted to various medical modalities such as pathology \cite{deng2025segment}. In addition, a few variants have attempted to reduce the generalization gap between natural and medical images through large-scale domain-specific pretraining and targeted fine-tuning. For instance, the MedSAM model is trained on more than 1.5M medical image--mask pairs spanning 10 modalities and 30+ anatomies \cite{ma2024segment}. Similarly, SAM-Med2D scales further by fine-tuning on nearly 4.6M images with 19.7M masks and adapting prompts (points/boxes/masks) and encoder--decoder weights for medical 2D data \cite{cheng2023sammed2d}. In addition, recent extensions such as MedSAM2 build on SAM2 to support prompt segmentation in 3D volumes and videos, narrowing the gap for volumetric CT/MRI and endoscopic/ultrasound sequences \cite{ma2025medsam2}. Beyond segmentation, medical vision--language models enable reporting, visual question-answering (VQA), grounding, and retrieval on biomedical images. Prominent examples include XrayGPT that aligns a medical visual encoder with a fine-tuned LLM to converse about chest radiographs and summarize reports \cite{thawakar2024xraygpt}. Similarly, LLaVA-Med trains a conversational assistant on PubMed Central figure--caption data using GPT-4--generated instructions, thereby demonstrating strong biomedical VQA capabilities \cite{li2023llava}. In the context of medical AIFMs, cross-modality generalization is essential, i.e., models should reason over heterogeneous imaging sources (and paired non-imaging data when available) to achieve greater robustness and diagnostic accuracy \cite{zhang2024challenges,ji2024segment}.

\subsubsection{Applications in Healthcare Delivery}
LLMs can significantly enhance healthcare delivery by streamlining patient communication, generating clear materials, and powering telemedicine platforms \cite{dash2023evaluation}. For instance, the literature demonstrates that they can improve operational efficiency by automating administrative tasks and supporting clinical decision-making for dental medicine \cite{eggmann2023implications}. LLMs play a key role in quality improvement, health education, and remote patient monitoring, fostering proactive interventions \cite{vaishya2023chatgpt}. Moreover, LLMs can address language barriers, facilitate care coordination, and collectively contribute to more effective, patient-centered healthcare delivery.

\subsubsection{Applications in Medical Education}
LLMs significantly contribute to the evolution of medical education \cite{eysenbach2023role,kung2023performance}. LLMs have numerous applications in medical education; for instance, they can facilitate curriculum development, generate diverse educational content \cite{van2023if}, support global accessibility through language translation \cite{fraiwan2023review}, and assist with assessment and evaluation using adaptive tools \cite{abd2023large}. Furthermore, LLM- and AIFM-empowered simulators can enable lifelike clinical case simulations, support interactive e-learning platforms \cite{teebagy2023improved}, contribute to continuing medical education \cite{antaki2023evaluating}, and facilitate the summarization of intricate medical literature \cite{gill2023chatgpt}. Also, LLMs have the potential to significantly impact and transform psychological health assessment education \cite{kjell2023ai}. These applications improve the efficiency, accessibility, and personalization of medical education, meeting the dynamic requirements of students and professionals in the healthcare sector \cite{li2024chatgpt,ray2023chatgpt}. In practical terms, the Chatdoctor \cite{li2023chatdoctor} and other LLMs could be designed to assist users with medical queries, provide information about symptoms, offer general health advice, or even engage in conversation about medical topics \cite{wu2024pmc,haupt2023ai,javaid2023chatgpt,moons2023chatgpt}. Moreover, some studies suggest that ChatGPT can provide medical information about symptoms, diseases, and medications \cite{iftikhar2023docgpt,muftic2023exploring}, although its accuracy and currency remain contested.

\subsubsection{Applications in Telehealth}
LLMs offer a range of applications in telehealth that can significantly enhance the delivery and accessibility of healthcare services. For instance, LLM-based virtual assistance and chatbots can provide patients with timely information and support. Similarly, LLMs can enable healthcare providers to efficiently assess patient conditions and prioritize care based on urgency. In addition, LLMs can streamline appointment scheduling and send automated reminders, thereby improving patient compliance and reducing no-show rates. They also play a critical role in patient health monitoring and reporting, enabling continuous assessment of health metrics and timely intervention when necessary. Furthermore, LLMs can assist in remote follow-ups, providing patients with essential information regarding their treatment plans and recovery processes. The integration of LLMs in telehealth not only enhances the efficiency of healthcare delivery but also improves patient outcomes by ensuring individuals receive personalized and timely care \cite{snoswell2023artificial}.

\subsubsection{Applications in Surgical Science}
AI models, particularly Convolutional Neural Networks (CNNs) and Vision Transformers (ViTs), have demonstrated strong performance in medical image recognition tasks, including intraoperative and preoperative surgical analysis. These models can identify anatomical structures, detect anomalies, and assist in surgical scene segmentation, thereby enhancing diagnostic precision and aiding in surgical planning \cite{ahmed2024deep}. In robot-assisted surgery, AI algorithms process real-time visual data to guide robotic arms, optimize instrument trajectories, and reduce invasiveness while maintaining procedural accuracy. Moreover, AI-powered predictive maintenance systems can forecast when surgical instruments require calibration or replacement, ensuring optimal equipment performance during operations. Such proactive, data-driven approaches contribute to safer, more efficient, and precision-guided surgical workflows.

Across these application areas, several recurring computational themes emerge. First, multimodal integration is becoming increasingly important, particularly in applications that combine medical images, clinical notes, and structured EHR data for diagnosis, reporting, and surgical support. Second, grounding models in external evidence---through retrieval, image--report alignment, or patient-specific clinical context---can help constrain generation when parametric knowledge alone is insufficient. Third, reasoning and instruction-following provide a common interface for applications such as clinical question answering, conversational VQA, decision support, and medical education. Finally, the growing use of these models in clinically consequential settings highlights shared concerns regarding reliability, privacy, bias, and appropriate human oversight. The case study presented later in this paper takes up the second of these themes directly, evaluating whether retrieval-based grounding in the Summary of Product Characteristics improves open-weight and proprietary LLMs on a prescription-suitability task.

\begin{table*}
\centering
\scriptsize
\caption{Open AIFMs for healthcare applications.
\textbf{W}: weights (\cmark~public, \gmark~gated, \xmark~not released).
\textbf{license}: weight licensing (\osi~OSI-approved, \lcl~community/use-restricted, \ncl~non-commercial, \ndl~non-commercial/no-derivatives, \bsp~bespoke terms, \nol~none declared, \nal~not applicable).
\textbf{C}: training code; \textbf{D}: training data; \pmark~partial or access-controlled release.
\textbf{OSAID}: compliance with the Open Source AI Definition v1.0~\cite{osi2024osaid} and ``n/s'' denotes not stated.}

\label{tab:OpenAI-models}
\begin{tabular}{p{4.0cm} p{2.6cm} p{1.5cm} c c c c c}
\toprule
\textbf{Model (year)} & \textbf{Released params} & \textbf{Ref.} &
\textbf{W} & \textbf{license} & \textbf{C} & \textbf{D} & \textbf{OSAID} \\
\midrule

\multicolumn{8}{c}{\textit{Biomedical text encoders and generators}} \\
\cmidrule(lr){1-8}
BioBERT (2020)            & 110M, 340M       & \cite{lee2020biobert}      & \cmark & \nol  & \pmark & \xmark & \xmark \\
BioMegatron (2020)        & 345M             & \cite{shin2020biomegatron} & \cmark & \bsp  & \cmark & \xmark & \xmark \\
BioELECTRA (2021)         & 110M             & \cite{raj2021bioelectra}   & \cmark & \nol  & \xmark & \xmark & \xmark \\
KeBioLM (2021)            & n/s              & \cite{yuan2021improving}   & \cmark & \nol  & \pmark & \xmark & \xmark \\
BioBART (2022)            & 140M, 406M       & \cite{yuan2022biobart}     & \cmark & \osi  & \cmark & \xmark & \xmark \\
BioLinkBERT (2022)        & 110M, 340M       & \cite{yasunaga2022linkbert}& \cmark & \osi  & \pmark & \xmark & \xmark \\
BioGPT (2022)             & 347M, $\sim$1.5B & \cite{luo2022biogpt}       & \cmark & \osi  & \pmark & \xmark & \xmark \\
RadBERT (2022)            & n/s              & \cite{yan2022radbert}      & \cmark & \osi  & \xmark & \xmark & \xmark \\
\midrule

\multicolumn{8}{c}{\textit{Clinical and EHR models}} \\
\cmidrule(lr){1-8}
ClinicalBERT (2019)       & n/s              & \cite{huang2019clinicalbert}& \cmark & \nol & \cmark & \pmark & \xmark \\
Bio\_ClinicalBERT (2019)  & 110M             & \cite{alsentzer2019publicly}& \cmark & \osi & \pmark & \pmark & \xmark \\
Med-BERT (2021)           & 17M (withheld)   & \cite{rasmy2021med}        & \xmark & \nal   & \cmark & \xmark & \xmark \\
CPLLM (2024)              & None declared   & \cite{ben2024cpllm}        & \xmark & \nol  & \cmark & \xmark & \xmark \\
\midrule

\multicolumn{8}{c}{\textit{Medical dialogue and question answering}} \\
\cmidrule(lr){1-8}
ChatDoctor (2023)         & 7B (LLaMA)       & \cite{li2023chatdoctor}    & \cmark & \ncl  & \cmark & \cmark & \xmark \\
DoctorGLM (2023)          & 6B (adapters)    & \cite{xiong2023doctorglm}  & \pmark & \nol  & \cmark & \pmark & \xmark \\
HuatuoGPT-II (2024)       & 7B, 13B, 34B     & \cite{chen2024huatuogpt2}  & \cmark & \osi  & \cmark & \cmark & \pmark$^{\dag}$ \\
Meditron (2023)           & 7B, 70B          & \cite{chen2023meditron}    & \gmark & \lcl  & \cmark & \pmark & \xmark \\
Meditron-3 (2025)       & 8B, 70B          & \cite{sallinen2025llama}                 & \gmark & \lcl  & \pmark & \xmark & \xmark \\
Med42-v2 (2024)           & 8B, 70B          & \cite{christophe2024med42} & \cmark & \lcl  & \xmark & \xmark & \xmark \\
OpenBioLLM (2024)         & 8B, 70B          & \cite{OpenBioLLMs}                & \cmark & \lcl  & \xmark & \xmark & \xmark \\
Bio-Medical-Llama-3 (2024)& 8B               & \cite{ContactDoctor_Bio-Medical-Llama-3-8B}                & \gmark & \ncl  & \xmark & \xmark & \xmark \\
MedGemma 27B text (2025)  & 27B              & \cite{sellergren2025medgemma}& \gmark & \bsp & \xmark & \xmark & \xmark \\
\midrule

\multicolumn{8}{c}{\textit{Medical reasoning models}} \\
\cmidrule(lr){1-8}
HuatuoGPT-o1 (2024)       & 7B, 8B, 70B, 72B     & \cite{chen2024huatuogpt}    & \cmark & \osi  & \pmark & \pmark & \xmark \\
MedReason (2025)          & 8B               & \cite{wu2025medreason}     & \cmark & \osi  & \pmark & \pmark & \xmark \\
m1 (2025)                 & 7B, 32B              & \cite{huang2025m1}         & \cmark & \osi  & \pmark & \pmark & \xmark \\
Baichuan-M2 (2025)        & 32B              & \cite{dou2025baichuan}      & \cmark & \osi  & \xmark & \xmark & \xmark \\
\midrule

\multicolumn{8}{c}{\textit{Medical vision--language models}} \\
\cmidrule(lr){1-8}
MedGemma 4B (2025)        & 4B             & \cite{sellergren2025medgemma}& \gmark & \bsp & \xmark & \xmark & \xmark \\
MedGemma 1.5 4B (2026)    & 4B             & \cite{sellergren2026medgemma}                 & \gmark & \bsp  & \xmark & \xmark & \xmark \\
MedSigLIP-448 (2025)      & 800M             & \cite{sellergren2025medgemma}& \gmark & \bsp & \xmark & \xmark & \xmark \\
Lingshu (2026)            & 7B, 32B          & \cite{xu2026lingshu}     & \cmark & \osi  & \pmark & \pmark & \xmark \\
HealthGPT (2025)          & 3.8B, 14B              & \cite{lin2025healthgpt}    & \cmark & \osi  & \pmark & \pmark & \xmark \\
Dragonfly-Med v2 (2024)   & 8B               & \cite{thapa2024dragonfly}  & \gmark & \ncl  & \cmark & \pmark & \xmark \\
MAIRA-2 (2024)            & 7B            & \cite{bannur2024maira}    & \gmark & \bsp  & \xmark & \pmark & \xmark \\
Merlin (2026)             & n/s              & \cite{blankemeier2026merlin}& \cmark & \osi & \cmark & \pmark & \xmark \\
\midrule

\multicolumn{8}{c}{\textit{Pathology and imaging encoders}} \\
\cmidrule(lr){1-8}
CONCH (2024)              & ---              & \cite{lu2024visual}         & \gmark & \ndl  & \pmark & \xmark & \xmark \\
UNI2-h (2025)$^\S$             & 681M              & \cite{chen2024towards}         & \gmark & \ndl  & \pmark & \xmark & \xmark \\
Virchow2 (2024)           & 632M             & \cite{zimmermann2024virchow2}& \gmark & \ndl & \xmark & \xmark & \xmark \\
Prov-GigaPath (2024)      & ---              & \cite{xu2024whole}      & \gmark & \osi  & \pmark & \xmark & \xmark \\
TITAN (2025)              & 159M             & \cite{ding2025multimodal}       & \gmark & \ndl  & \pmark & \xmark & \xmark \\
\midrule

\multicolumn{8}{c}{\textit{Medical speech}} \\
\cmidrule(lr){1-8}
MedASR (2026)             & 105M             & \cite{wu2026medasr}        & \gmark & \bsp  & \pmark & \xmark & \xmark \\
\midrule

\multicolumn{8}{c}{\textit{Therapeutics and biomolecular}} \\
\cmidrule(lr){1-8}
TxGemma (2025)            & 2B, 9B, 27B      & \cite{wang2025txgemma}     & \gmark & \bsp  & \xmark & \xmark & \xmark \\
\midrule

\multicolumn{8}{c}{\textit{General-purpose backbones used in healthcare}} \\
\cmidrule(lr){1-8}
Llama 2 (2023)            & 7B, 13B, 70B     & \cite{llama2}   & \gmark & \lcl  & \xmark & \xmark & \xmark \\
Chameleon (2024)          & 7B, 30B$^{\ddag}$& \cite{team2024chameleon}   & \gmark & \ncl  & \xmark & \xmark & \xmark \\
Llama 4 Scout (2025)      & 109B (17B act.)  & \cite{meta2025llama4}                 & \gmark & \lcl  & \xmark & \xmark & \xmark \\
Mistral Small 3.2 (2025)  & 24B              & \cite{mistral2025small32}                 & \cmark & \osi  & \xmark & \xmark & \xmark \\
gpt-oss (2025)            & 21B, 117B        & \cite{agarwal2025gpt}    & \cmark & \osi  & \xmark & \xmark & \xmark \\
\bottomrule
\end{tabular}

\vspace{2pt}
{\footnotesize
$^{\dag}$~HuatuoGPT-II releases weights under Apache-2.0 together with its
training code and its pretraining and instruction-tuning corpora; its base
models (Baichuan2, Yi) carry separate upstream terms; therefore, full OSAID compliance
is unresolved rather than established.
$^{\ddag}$~The released Chameleon checkpoints accept mixed-modal input, but
produce text only; the image-generation model was withheld from the release.
$^\S$~UNI2-h was released in January 2025; its citation refers to the original UNI publication.
}
\end{table*}

\begin{table*}
\centering
\scriptsize
\caption{Publicly documented healthcare datasets used for LLM and AIFM
development and evaluation, grouped by access tier. \textbf{Open}: anonymous
download. \textbf{Registered}: free account or click-through agreement.
\textbf{Credentialed}: identity verification, human-subjects training, and a
signed data use agreement. \textbf{Controlled}: approval by a data access
committee.}
\label{tab:open_source_dataset}
\begin{tabular}{p{2.4cm} p{1.9cm} p{2.0cm} p{5.6cm} p{1.5cm} p{2.0cm}}
\toprule
\textbf{Dataset} & \textbf{Modality} & \textbf{Task type} &
\textbf{Size} & \textbf{Access} & \textbf{License} \\
\midrule

\multicolumn{6}{c}{\textit{Open access}} \\
\cmidrule(lr){1-6}
Observed Antibody Space \cite{olsen2022observed} & Antibody sequences & Protein modeling &
1.5B unpaired sequences from 80 studies and paired sequences from 5 studies & Open & CC-BY-4.0 \\

Big Fantastic Database (BFD) \cite{steinegger2018clustering,jumper2021highly} & Protein sequences & Protein folding, MSA &
Clustered from 2.5B sequences & Open & CC-BY-4.0 \\

RNAcentral \cite{rnacentral2026rnacentral} & ncRNA sequences & Structure prediction &
$>$45M sequences, $>$30M secondary structures (Release 26) & Open & CC0 \\

ZINC20 \cite{irwin2020zinc20} & Chemical compounds & Molecule generation &
1.4B compounds, 310 catalogs, 150 companies & Open & Free-use terms \\

ZINC-22 \cite{tingle2023zinc} & Chemical compounds & Molecule generation &
$>$37B compounds, $>$4.5B ready-to-dock 3D compounds &
Open & Free-use terms \\

MedMNIST v2 \cite{yang2023medmnist} & 2D and 3D images & Classification &
18 datasets (12$\times$2D, 6$\times$3D); 708,069 2D + 9,998 3D images & Open & CC-BY-4.0 \\

MedQA (USMLE) \cite{jin2021disease} & Medical QA & Clinical reasoning &
12,723 questions (train 10,178 / dev 1,272 / test 1,273) & Open & MIT \\

PubMedQA \cite{jin2019pubmedqa} & Biomedical QA & Evidence-based QA &
Contains 1,000 expert-labelled, 61.2K unlabelled, and 211.3K artificially generated QA instances & Open & MIT \\

cMedQA2 \cite{8548603} & Chinese medical QA & Answer selection & 108,000 questions, 203,569 answers & Open & GPL-3.0 \\

Huatuo-26M \cite{wang2025huatuo} & Chinese medical QA & Instruction tuning &
26.5M QA pairs; consultation records provide answers as URLs & Open & Apache-2.0 \\

NHANES \cite{cdc2024nhanes} & Health surveys & Epidemiology &
Continuous 2-year cycles since 1999 & Open $+$ Controlled & NCHS Data User Agreement \\

\midrule
\multicolumn{6}{c}{\textit{Registered access}} \\
\cmidrule(lr){1-6}
CT-RATE \cite{hamamci2026generalist} & 3D chest CT $+$ reports & Vision--language &
25,692 3D CT volumes from 21,304 patients & Registered & CC-BY-NC-SA-4.0 \\

MedTrinity-25M \cite{xie2025medtrinity} & Multimodal images & Instruction tuning &
$>$25M images, 10 modalities, $>$65 diseases & Registered & None declared \\

BIOMEDICA \cite{lozano2025biomedica} & Article figures $+$ text & Vision--language pretraining &
$>$24M image--text pairs from $>$6M PubMed Central open access articles & Registered & None declared \\

\midrule
\multicolumn{6}{c}{\textit{Credentialed access}} \\
\cmidrule(lr){1-6}
MIMIC-III \cite{johnson2016mimic} & ICU EHR & Outcome prediction &
53,423 adult patient admissions ($\geq$16 y) $+$ 7,870 neonatal & Credentialed & PhysioNet CHD 1.5.0 \\

MIMIC-IV \cite{johnson2023mimic} & Hospital $+$ ICU EHR & Clinical event modeling &
364,627 patients; 546,028 hospitalizations; 94,458 ICU stays & Credentialed & PhysioNet CHD 1.5.0 \\

MIMIC-CXR \cite{johnson2019mimic} & Chest radiographs $+$ reports & Vision--language &
377,110 images, 227,835 studies, 65,379 patients & Credentialed & PhysioNet CHD 1.5.0 \\

\midrule
\multicolumn{6}{c}{\textit{Split open and controlled}} \\
\cmidrule(lr){1-6}
TCGA \cite{NCI_TCGA} & Genomics, pathology & Multimodal profiling &
$>$20,000 primary cancer and matched normal samples spanning 33 cancer types &
Open $+$ Controlled  & NIH GDS / dbGaP \\

\bottomrule
\end{tabular}

\end{table*}

Table \ref{tab:OpenAI-models} catalogues healthcare and healthcare-adjacent FMs, scoring each on four release-openness axes: weight availability and gating, the license declared on the weights, and whether training code and training data were released. Each model is also scored against the Open Source AI Definition v1.0 \cite{osi2024osaid}. The listed models span a range of healthcare NLP tasks, including named entity recognition (NER), clinical text classification, and relation extraction. They draw on domain-specific pretraining corpora such as PubMed abstracts, clinical notes from MIMIC-III, and structured EHR data. BioBERT \cite{lee2020biobert}, ClinicalBERT \cite{huang2019clinicalbert}, Med-BERT \cite{rasmy2021med}, and ChatDoctor \cite{li2023chatdoctor} are representative examples; He et al. \cite{he2025survey} provide a broader review of LLMs developed for the healthcare domain. The general-purpose backbones in Table \ref{tab:OpenAI-models} are included because they are widely used as base models for healthcare adaptation (e.g., Llama 2 underlies Meditron; Llama 3 underlies Med42-v2) and also, because we evaluate them in our case study (Llama 4 Scout, Mistral Small 3.2 in Section~\ref{sec:case_study}). Table~\ref{tab:open_source_dataset} lists the datasets used to train and evaluate these models, grouped by access tier---from openly downloadable resources such as PubMedQA and MedQA to credentialed clinical records (MIMIC-III, MIMIC-IV) and controlled-access genomic data (TCGA)---and annotated with their licence terms.

\subsection{Model Identification and Openness Scoring}
\label{sec:protocol}
We identified candidate models through three sources: (i) published articles and preprints; (ii) open model repositories (Hugging Face Hub, GitHub); and (iii) model indexes maintained by major releasing organizations. The search covered 2019 to September 2026. A model was included if it (a) is presented by its developers as a foundation model or large language model, (b) targets a health, clinical, or biomedical application, or is a general-purpose backbone on which the healthcare models in
this survey are built, and (c) has a citable release artifact: a peer-reviewed paper, a preprint, or a versioned model card. For every included model we report four release properties: weight availability, access gating, the license declared on the weights, and whether training code and training data were released. Where a repository's code license and its weights license differ, we record the license attached to the weights, since that is the instrument that governs reuse. We score each model against OSAID \cite{osi2024osaid}, which requires three components: sufficiently detailed information about the training data that a skilled person could build a substantially equivalent system; the complete training and inference source code; and model parameters made available under OSI-approved terms.

\subsection{Safety Considerations of Open AI for Healthcare}

\subsubsection{Safety in Pretraining}
Safety in pretraining refers to measures and considerations that ensure the ethical, responsible, and secure development of healthcare LLMs during the initial training phase. Pretraining involves training a language model on a large, diverse text corpus before fine-tuning it for specific tasks. Ensuring safety during this early phase is crucial to prevent the propagation of biases, misinformation, or harmful content in the subsequent fine-tuning and deployment stages \cite{rando2023universal}. This will ensure sensitive information is protected from being compromised, provide accurate and reliable medical information, reduce bias in outputs for fair recommendations, and maintain interpretability and transparency in decisions.

\subsubsection{Safety in Fine-Tuning}
Safety during the fine-tuning of healthcare LLMs is a critical aspect of developing reliable and trustworthy AI systems for medical applications. Fine-tuning involves training the model on domain-specific data to enhance its performance in healthcare tasks. Ensuring safety during fine-tuning involves protecting patient privacy, ethically handling data, prioritizing clinical accuracy, mitigating bias, providing explainability, testing robustness, conducting regular audits, and collaborating with healthcare professionals to create reliable, accurate, and ethical models for real-world medical applications \cite{291199}. Below, we briefly describe techniques that can be used to ensure the safety of LLMs in healthcare applications during the fine-tuning process.

\begin{itemize}
    \item \textit{Supervised Safety Fine-Tuning}: It incorporates adversarial prompts and safe demonstrations into the initial supervised fine-tuning phase. This approach instills adherence to safety guidelines from the outset, forming the basis for subsequent high-quality human preference data annotation \cite{llama2}.
    \item \textit{Safety using Reinforcement Learning from Human Feedback (RLHF)}: This technique involves training a safety-specific reward model and collecting more challenging adversarial prompts for rejection sampling-style fine-tuning and Proximal Policy Optimization (PPO) optimization \cite{llama2}. Throughout the RLHF stage, accumulating iterative reward-modeling data in parallel with model enhancements is crucial to ensure that the reward models remain within the distribution.
    \item \textit{Safety via Instruction/Context Distillation}: In this approach, a smaller or open model is fine-tuned on safety-aligned demonstrations produced by a stronger teacher or by a constitutional prompting framework, so that safety behaviors (e.g., refusal of hazardous instructions, adherence to ethical constraints) are internalized without requiring safety system prompts at inference. This is closely related to the Constitutional AI method \cite{bai2022constitutional} and knowledge distillation for safety \cite{snell2022learning}, and can be seen as applying context distillation to the safety domain.
\end{itemize}

Fine-tuning clinical LLM models involves adapting them to process clinical and biomedical texts effectively, as well as other medical data, such as medical images, pathology reports, and diagnostic test results. This strategic approach enables the model to derive insights from diverse data modalities, providing a more comprehensive view of patient information and enhancing the performance of healthcare-related tasks.

\subsection{What ``Open'' Turns Out to Mean}
\label{sec:openness_audit}

Table \ref{tab:openness_audit} summarizes the release metadata of the 45 models in Table~\ref{tab:OpenAI-models}. The data reveal four patterns that a binary open/closed distinction cannot express. \textit{First, only one in three of these models is unrestricted.} Of the 45 models, 35.6\% carry an OSI-approved license and 51.1\% are ungated, but only 15 (33.3\%) are both. A reader who assumes that a model described in the literature as open-source can be downloaded and reused without conditions will be wrong more often than right. No model in the set satisfies all three criteria of the Open Source AI Definition \cite{osi2024osaid}: models with OSI-approved licenses release neither training code nor the training-data information the definition requires, while models that do release code and data attach restrictive licenses to the weights. The closest case is HuatuoGPT-II, which publishes Apache-2.0-Licensed weights alongside its training code and corpora but inherits separate terms from its base models (Baichuan2 and Yi). \textit{Second, license type and access gating reinforce each other.} Seven of the eight models released under bespoke terms of use are gated, as are all four released under CC-BY-NC-ND. Community-licensed models split four gated into two ungated. The pattern inverts for undocumented releases: of the six repositories that declare no license at all, four are public, one is partial, and one withholds weights. Yet, gating is not a proxy for restriction: Prov-GigaPath is Apache-2.0 yet gated, and its access form imposes a research-use condition that the license itself does not carry, while Merlin is MIT-licensed with no gate.

\begin{table}
\centering
\caption{Release openness of the 45 healthcare and healthcare-adjacent
foundation models in Table \ref{tab:OpenAI-models}.}
\label{tab:openness_audit}
\begin{tabular}{lrr}
\toprule
\textbf{License class} & \textbf{Models} & \textbf{Share} \\
\midrule
OSI-approved (Apache-2.0, MIT)      & 16 & 35.6\% \\
Bespoke terms of use                &  8 & 17.8\% \\
Community license (use-restricted)  &  6 & 13.3\% \\
Non-commercial                      &  4 &  8.9\% \\
Non-commercial, no derivatives      &  4 &  8.9\% \\
None declared                       &  6 & 13.3\% \\
Not applicable (weights withheld)   &  1 &  2.2\% \\
\midrule
\textbf{Not OSI-approved}           & \textbf{29} & \textbf{64.4\%} \\
\textbf{Gated}                      & \textbf{19} & \textbf{42.2\%} \\
\textbf{Ungated and OSI-approved}   & \textbf{15} & \textbf{33.3\%} \\
\bottomrule
\end{tabular}
\end{table}

\textit{Third, declared licenses do not always reflect effective reuse rights.} HuatuoGPT-o1-8B and its derivative MedReason-8B both carry Apache-2.0, yet both descend from Llama 3.1-8B-Instruct, whose community license imposes additional terms on redistribution. HuatuoGPT-o1's own ethical statement goes on to restrict clinical and commercial use, despite the permissive license on the weights. In pathology, the pattern runs in the opposite direction: Paige released the original Virchow under Apache-2.0 but tightened Virchow2 to CC-BY-NC-ND-4.0. Openness, it turns out, is a per-release decision, not a stable property of a model line. \textit{Fourth, and most consequential for clinical adaptation, the dominant license in computational pathology forbids the operation that defines a FM.} CONCH, UNI2-h, Virchow2, and TITAN are released under CC-BY-NC-ND-4.0. The \textit{NoDerivatives} term prohibits distributing adapted versions of the work, yet fine-tuning a FM on a local cohort and releasing the result is the central use case that motivates open release in healthcare. CONCH's download terms go further, prohibiting commercial use of models trained on its outputs. These findings point to the same conclusion. The openness question in healthcare AI has been asked at the wrong granularity. What matters is not whether a model is open, but which of weights, gate, license, training code, and training data a developer has chosen to open, and whether that combination supports the downstream use for which the release is justified. Jewitt et al. \cite{jewitt2026permissive} documented license-integrity failures across the general AI supply chain, and Liesenfeld and Dingemanse \cite{liesenfeld2024rethinking} argued that openness in generative AI is necessarily composite and gradient.

\section{Leveraging Open LLMs for Prescription: A Case Study}
\label{sec:case_study}
This section presents a case study to investigate the prospects of using open LLMs in critical healthcare technologies, specifically for digitizing prescription analysis tasks. Our objective is to assess whether a proposed medication is suitable for a given patient profile, and to identify the patient-specific risks---including contraindications, interactions, and dosing concerns---that bear on that judgment. Our motivation to focus on this topic is inspired by the work of Pais et al. \cite{pais2024large}, which leverages LLMs to enhance the accuracy of medication directions in pharmacy operations. While their work emphasizes ensuring the correct administration of medications, our study evaluates the suitability of the medication in the first place, i.e., determining which medicine product is relatively safer for a particular patient. Our investigation assesses the feasibility of using open LLMs for critical prescription tasks and compares their performance with proprietary counterparts. This case study also allows us to address the adaptation of open LLMs to accurately contextualize complex patient and drug information for actionable insights that guide safe prescribing practices. Specifically, we aim to identify performance gaps and understand the viability of using open LLMs in complex healthcare contexts. We seek to answer the following pressing research questions:

\begin{itemize}
    \item[\textbf{RQ1}:] How accurately do open LLMs assess medication suitability and flag patient-specific prescribing risks, compared with a proprietary baseline?
    \item[\textbf{RQ2}:] What are the limitations and challenges of using open LLMs in prescription analysis?
    \item[\textbf{RQ3}:] How does retrieval-augmented grounding affect open and proprietary LLMs on prescription suitability, and does it close or widen the gap between them?
\end{itemize}

\subsection{Defining the Problem}
Safer prescribing practices are essential but increasingly challenging due to the vast amount of clinical guidance and patient-specific data that must be reviewed to make optimal decisions. Manual exploration of this information is impractical, given limited consultation times. LLMs offer a solution to this challenge. In this paper, we present a case study on using LLMs to enhance prescribing safety by assessing the suitability of a proposed medication for an individual patient given a structured patient profile. Each profile is synthetic and records demographics (age, sex, ethnicity), a primary diagnosis, one proposed medication and dose, comorbidities, allergies, current medications, pregnancy, lactation, and a named risk scenario. To assess adverse drug reactions (ADRs) and interactions at the point of prescribing, we use a customized prompt that includes the patient profile and the Summary of Product Characteristics (SmPC) for the given medicine. The LLM is tasked with identifying the suitability of the medication by highlighting various classes of interactions, including (1) Age; (2) Comorbidities; (3) Contraindications; (4) Dose; (5) Genetics; (6) Lactation; (7) Pregnancy; and (8) Warnings. The SmPC is an essential regulatory document used in Europe and the United Kingdom for medicinal products, providing healthcare professionals with crucial information for the safe and effective administration of a drug. It serves as a comprehensive source of information, covering therapeutic indications, dosage, administration, contraindications, and side effects \cite{EU_SmPC}. For the purposes of this study we group the SmPC into the six blocks described below. 

\begin{enumerate}
    \item \textit{Composition:} Active ingredients and excipients.
    \item \textit{Pharmaceutical Form:} The physical form of the drug, such as a tablet or injection.
    \item \textit{Clinical Particulars:} Information on indications, dosage, contraindications, and precautions.
    \item \textit{Pharmacological Properties:} Details on the mechanism of action, absorption, metabolism, and other pharmacokinetic aspects.
    \item \textit{Pharmaceutical Particulars:} Guidance on storage conditions, handling requirements, and shelf life.
    \item \textit{Regulatory and Administrative Information:} Legal classification and marketing authorization details (where applicable).
\end{enumerate}

This information should be reviewed alongside the patient profile to support safer prescribing and prevent ADRs.

\begin{figure*}
\centering
\includegraphics[width=\linewidth]{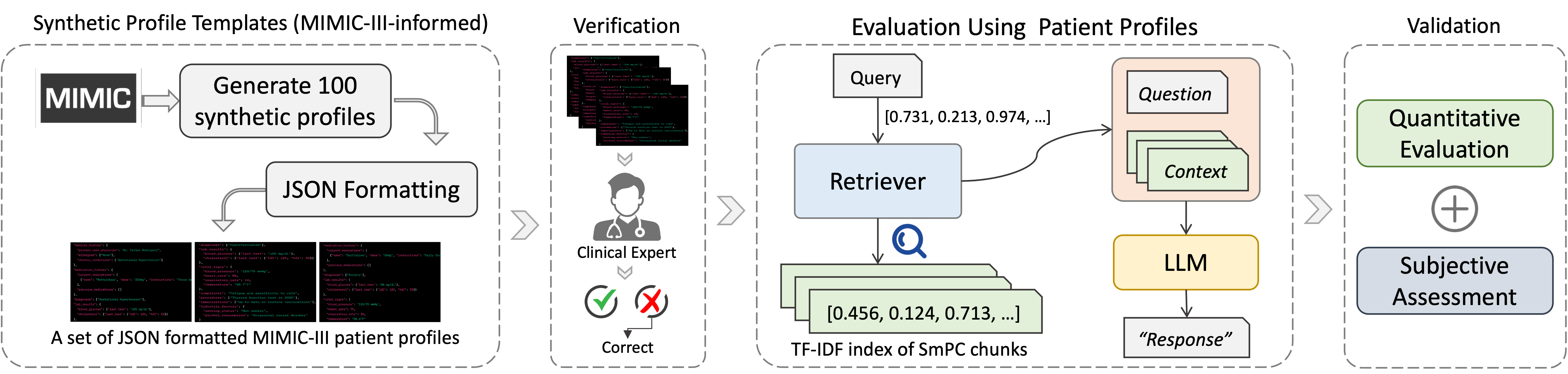}
\caption{Illustration of the proposed methodology for evaluating open-weight and proprietary LLMs using two settings, i.e., with and without Retrieval-Augmented Generation (RAG).}
\label{fig:method}
\end{figure*}

\subsection{Proposed Method for Prescription Evaluation}
To evaluate the proficiency of open and proprietary LLMs in encoding clinical knowledge and their potential for personalized prescriptions, we conducted a controlled experimental study. Specifically, we examined the effectiveness of various LLMs, including both open-weight and proprietary models. As shown in Figure \ref{fig:method}, our proposed methodology consists of four major steps: construction of synthetic patient profiles from MIMIC-informed clinical templates, clinician verification of those profiles, LLM evaluation on the prescription-suitability task, and quantitative plus qualitative validation. Below, we provide a brief description of each step:

\subsubsection{Data Description} 

We constructed a dataset of 100 unique synthetic patient profiles using clinical variables and treatment patterns documented in the Medical Information Mart for Intensive Care (MIMIC)-III database \cite{johnson2016mimic}. Note that no real patient records or patient-level data were reproduced. Instead, MIMIC-III informed the selection of demographic attributes, diagnoses, medications, and clinically relevant risk factors represented in the templates. Our dataset is evenly distributed across five medication strata: warfarin, metformin, levothyroxine, lisinopril, and omeprazole, with 20 profiles assigned to each stratum. Within every stratum, the profiles represent five predefined scenarios comprising a baseline case and medication-specific risk conditions, such as pregnancy, renal impairment, contraindicating comorbidities, and potentially interacting treatments.

Each profile was encoded as a structured JSON record designed to evaluate whether a model could assess the suitability of a proposed medication and provide field-level reasoning grounded in patient-specific information. Each record contains a synthetic profile identifier, age, sex, ethnicity, primary diagnosis, proposed medication and dosage, comorbidities, allergies, current medications, pregnancy status, lactation status, and a predefined risk scenario. The primary diagnoses that correspond to the five medication strata are rheumatic heart disease, type 2 diabetes, hypothyroidism, hypertension, and gastric ulcer. To ensure that medication-interaction assessment was not uniformly trivial, a subset of profiles also included potentially interacting concomitant medications. The resulting dataset was developed exclusively for research evaluation and is not intended to support clinical decision-making or patient care.

\subsubsection{Verification of Patient Profiles} 

To ensure realism and clinical relevance, a clinician co-author with more than ten years of experience in multimorbidity and polypharmacy reviewed and verified all 100 synthetic patient profiles. If any inaccuracies or inconsistencies were identified among the various factors of a patient profile (e.g., demographics, diagnoses, or treatments), these were manually corrected. This verification process is a critical step in ensuring the validity of the experimental study and the clinical plausibility of the profiles for the stated risk scenarios.

\begin{table*}
\centering
\caption{Medication Suitability Evaluation Prompt Template}
\scriptsize
\begin{tabular}{p{17.8cm}}
\toprule
\textbf{Context}  \\
\midrule
You are an experienced and helpful prescribing assistant named Charlie. Charlie can support prescribers to carry out relevant checks when prescribing medication based on the patient's medical profile and medical knowledge. You should tell  if the medication is suitable for the user based on the following patient profile and provide the result in JSON format: \\
\begin{quote}
\centering
$<$patient profile$>$
\end{quote} \\
\midrule
\textbf{Instructions} \\
\midrule
For each check, use ``Suitable'', ``Risky'', or ``N/A'', and provide a reason and evidence field. Cite the source filename and page when retrieved context is supplied. Return an overall result, a score from 0 (clearly unsuitable) to 100 (clearly suitable), and a reason. \\
\midrule
\textbf{Output Format (JSON)}  \\
\midrule
\begin{verbatim}
{
  "Age": {"result": <result>, "reason": <reason>},
  "Dose": {"result": <result>, "reason": <reason>},
  "Comorbidities": {"result": <result>, "reason": <reason>},
  "Contraindications": {"result": <result>, "reason": <reason>},
  "Pregnancy": {"result": <result>, "reason": <reason>},
  "Lactation": {"result": <result>, "reason": <reason>},
  "Warnings": {"result": <result>, "reason": <reason>},
  "Genetics": {"result": <result>, "reason": <reason>},
  "Overall Suitability": {"result": <result>, "score": <score>, "reason": <reason>}
}
\end{verbatim} \\
\bottomrule
\end{tabular}
\label{tab:prompt_template}
\end{table*}

\subsubsection{Evaluating LLMs for Prescription} 

For this case study, we evaluated a representative selection of open-weight and proprietary LLMs that varied in architecture, scale, and domain specialization. The open-weight models included two general-purpose LLMs, Llama 4 Scout and Mistral Small 3.2 24B Instruct, and one medical-domain LLM, MedGemma 27B Text IT, an instruction-tuned Gemma 3 variant further trained on medical text. We used GPT-5.6 Luna as the proprietary baseline. The models vary in scale and access pathways, influencing their depth of contextual reasoning and factual recall capabilities. We evaluated these models in prescription tasks using verified patient profiles and five commonly prescribed medications: Warfarin, Metformin, Levothyroxine, Lisinopril, and Omeprazole. For each medication, the evaluation relied on its corresponding SmPC. The LLMs were tasked with determining medication suitability based on multiple clinical factors, including patient age, comorbidities, contraindications, dosage, genetics, lactation, pregnancy status, and safety warnings. Both open-weight and proprietary models were evaluated to compare their performance in accuracy and reliability for clinical decision-making. Furthermore, to improve LLMs' ability to extract relevant information from SmPCs with greater precision, we used RAG, which is widely recognized for addressing inherent limitations of LLMs, such as hallucination \cite{ng2024rag,ansari2025lightweight}. Retrieved passages are selected by TF-IDF cosine similarity between the input query and the chunked SmPC text of the corresponding medication.

\subsubsection{Performance Evaluation}

The results generated by the LLMs were validated using a dual-validation approach that includes quantitative evaluation and subjective assessment through expert review. To quantitatively evaluate the model's performance against expert field labels derived from the corresponding SmPC, we use standard metrics, including accuracy, precision, recall, and F1-score. Overall suitability is scored by exact agreement between the model's \texttt{overall\_suitability} result (\texttt{Suitable} or \texttt{Risky}) and the expert overall label. Moreover, to assess clinical significance, two raters---a pharmacy student and a clinician co-author---independently scored a blinded subset of 96 responses (12 profiles $\times$ four models $\times$ two RAG conditions) for clinical soundness, safety, and contextual appropriateness. Each reviewer rates every response on a scale from 0 (inappropriate) to 5 (highly appropriate) while considering five dimensions: (1) Medication Selection Accuracy (MSA); (2) Drug Interaction Detection (DID); (3) Patient-Specific Dosage Adjustment (PSDA); (4) Prescription Safety Score (PSS); and (5) General Assessment (GA). Quadratic-weighted Cohen's kappa is computed per dimension on the independent ratings. Both raw reviewer scores are retained, and the mean of the two ratings is used for model-level summaries.

\subsection{Results and Discussions}

\subsubsection{Experimental Setup and Implementation Details}

We evaluate the performance of four LLMs---GPT-5.6 Luna, Llama 4 Scout, Mistral Small 3.2 24B Instruct, and MedGemma 27B Text IT for prescription tasks using a customized prompt template. GPT-5.6 Luna, Llama 4 Scout, and Mistral Small 3.2 24B Instruct were accessed via OpenRouter\footnote{\url{https://openrouter.ai/docs}}, while MedGemma 27B Text IT was hosted on a Hugging Face Inference Endpoint\footnote{\url{https://huggingface.co/docs/inference-endpoints}}. We use 100 verified synthetic patient profiles, each paired with one proposed medication, yielding 100 cases per model--condition pair, that is, 800 model outputs in the first run and 1{,}600 across the two runs. Each prompt includes the patient's demographics, comorbidities, allergies, and current medications, along with a query assessing the appropriateness of a specific prescription (Table \ref{tab:prompt_template} illustrates the prompt structure). The SmPCs for the selected medications were parsed, cleaned, and split into overlapping word chunks; a TF-IDF retriever ranked chunks by cosine similarity. Each run was performed twice for each model as part of the sensitivity analysis. RAG supplied the LLMs with the top-ranked SmPC passages.

\paragraph{Inference Configuration and Reproducibility}
All models used temperature 0 and a maximum output length of 1,800 tokens. The patient profiles, prompt schema, retrieval parameters, and SmPC sources were fixed across models and conditions. The primary results below use the first of the two runs, yielding one observation for each model--profile--condition combination. Model outputs were compared with the reference labels using exact agreement, except that \texttt{N/A} and \texttt{Suitable} were treated as equivalent for pregnancy, lactation, and genetics because both denote the absence of a profile-specific risk in these fields. Accuracy and macro-averaged precision, recall, and F1 were computed separately for each field, for overall suitability, and for each model and condition. Mean field accuracy is the unweighted mean of the eight field accuracies.

\begin{table*}
\centering
\scriptsize
\caption{Field-level and overall-suitability performance on 100 synthetic profiles using the first of two runs. P, R, and F1 are macro-averaged. Mean field accuracy is the unweighted mean of the eight field accuracies.}
\label{tab:results}
\begin{tabular*}{\textwidth}{@{\extracolsep{\fill}}llcccccccc}
\toprule
\multirow{2}{*}{\textbf{Model}} & \multirow{2}{*}{\textbf{Field}} &
\multicolumn{4}{c}{\textbf{No RAG}} & \multicolumn{4}{c}{\textbf{RAG}} \\
\cmidrule(lr){3-6} \cmidrule(lr){7-10}
& & \textbf{Acc.} & \textbf{P} & \textbf{R} & \textbf{F1} & \textbf{Acc.} & \textbf{P} & \textbf{R} & \textbf{F1} \\
\midrule
\multirow{9}{*}{GPT-5.6 Luna}
& Age & 0.880 & 0.872 & 0.813 & 0.836 & 0.810 & 0.586 & 0.479 & 0.517 \\
& Comorbidities & 0.940 & 0.961 & 0.897 & 0.922 & 0.860 & 0.611 & 0.533 & 0.562 \\
& Contraindications & 0.750 & 0.516 & 0.501 & 0.508 & 0.770 & 0.597 & 0.511 & 0.549 \\
& Dose & 0.640 & 0.736 & 0.709 & 0.638 & 0.820 & 0.600 & 0.553 & 0.572 \\
& Genetics & 0.950 & 0.500 & 0.475 & 0.487 & 0.980 & 0.500 & 0.490 & 0.495 \\
& Lactation & 1.000 & 1.000 & 1.000 & 1.000 & 1.000 & 1.000 & 1.000 & 1.000 \\
& Pregnancy & 0.930 & 0.731 & 0.963 & 0.796 & 0.910 & 0.687 & 0.874 & 0.738 \\
& Warnings & 0.470 & 0.730 & 0.518 & 0.350 & 0.440 & 0.341 & 0.316 & 0.261 \\
& Overall & 0.780 & 0.614 & 0.663 & 0.627 & 0.930 & 0.849 & 0.870 & 0.859 \\ \midrule
\multirow{9}{*}{Llama 4 Scout}
& Age & 0.760 & 0.739 & 0.579 & 0.575 & 0.760 & 0.876 & 0.556 & 0.529 \\
& Comorbidities & 0.910 & 0.926 & 0.855 & 0.882 & 0.910 & 0.633 & 0.563 & 0.588 \\
& Contraindications & 0.550 & 0.481 & 0.368 & 0.412 & 0.750 & 0.579 & 0.496 & 0.531 \\
& Dose & 0.620 & 0.531 & 0.509 & 0.446 & 0.670 & 0.640 & 0.610 & 0.611 \\
& Genetics & 0.970 & 0.500 & 0.485 & 0.492 & 0.990 & 0.500 & 0.495 & 0.497 \\
& Lactation & 1.000 & 1.000 & 1.000 & 1.000 & 0.990 & 0.500 & 0.495 & 0.497 \\
& Pregnancy & 0.920 & 0.714 & 0.957 & 0.778 & 0.910 & 0.687 & 0.874 & 0.738 \\
& Warnings & 0.550 & 0.419 & 0.387 & 0.348 & 0.450 & 0.313 & 0.317 & 0.285 \\
& Overall & 0.820 & 0.682 & 0.776 & 0.708 & 0.900 & 0.789 & 0.822 & 0.804 \\ \midrule
\multirow{9}{*}{MedGemma 27B}
& Age & 0.760 & 0.739 & 0.579 & 0.575 & 0.770 & 0.593 & 0.398 & 0.410 \\
& Comorbidities & 0.820 & 0.633 & 0.460 & 0.499 & 0.770 & 0.649 & 0.402 & 0.438 \\
& Contraindications & 0.350 & 0.490 & 0.229 & 0.310 & 0.520 & 0.628 & 0.337 & 0.369 \\
& Dose & 0.680 & 0.557 & 0.386 & 0.370 & 0.750 & 0.526 & 0.464 & 0.479 \\
& Genetics & 1.000 & 1.000 & 1.000 & 1.000 & 0.990 & 0.500 & 0.495 & 0.497 \\
& Lactation & 1.000 & 1.000 & 1.000 & 1.000 & 1.000 & 1.000 & 1.000 & 1.000 \\
& Pregnancy & 0.920 & 0.714 & 0.957 & 0.778 & 0.890 & 0.620 & 0.707 & 0.646 \\
& Warnings & 0.500 & 0.439 & 0.358 & 0.306 & 0.380 & 0.301 & 0.276 & 0.232 \\
& Overall & 0.850 & 0.713 & 0.793 & 0.740 & 0.930 & 0.862 & 0.840 & 0.850 \\ \midrule
\multirow{9}{*}{Mistral Small 3.2}
& Age & 0.750 & 0.872 & 0.537 & 0.496 & 0.810 & 0.572 & 0.448 & 0.469 \\
& Comorbidities & 0.900 & 0.592 & 0.593 & 0.592 & 0.710 & 0.511 & 0.388 & 0.435 \\
& Contraindications & 0.690 & 0.507 & 0.464 & 0.470 & 0.940 & 0.667 & 0.625 & 0.644 \\
& Dose & 0.670 & 0.828 & 0.554 & 0.494 & 0.720 & 0.846 & 0.622 & 0.605 \\
& Genetics & 0.960 & 0.500 & 0.480 & 0.490 & 0.990 & 0.500 & 0.495 & 0.497 \\
& Lactation & 1.000 & 1.000 & 1.000 & 1.000 & 0.990 & 0.500 & 0.495 & 0.497 \\
& Pregnancy & 0.920 & 0.671 & 0.723 & 0.693 & 0.890 & 0.661 & 0.863 & 0.707 \\
& Warnings & 0.570 & 0.693 & 0.605 & 0.531 & 0.440 & 0.336 & 0.318 & 0.260 \\
& Overall & 0.810 & 0.683 & 0.800 & 0.709 & 0.890 & 0.771 & 0.787 & 0.778 \\
\bottomrule
\end{tabular*}
\end{table*}

\subsubsection{Quantitative Results}

The main results of our experimental study on LLM-empowered personalized prescriptions are presented in Table \ref{tab:results}. The table reports the results in terms of different performance metrics, including accuracy, precision, recall, and F1-score for four different LLMs, namely, GPT-5.6 Luna, Llama 4 Scout, Mistral Small 3.2 24B Instruct, and MedGemma 27B Text IT. Moreover, the table provides a comparative analysis of all models across two settings: with and without augmenting LLM responses using RAG. Table \ref{tab:results} reveals that GPT-5.6 Luna attains the highest mean field-level accuracy without RAG (0.820), with Mistral Small 3.2 close behind (0.808). For overall prescription suitability, the open-weight point estimates exceed the proprietary baseline without RAG: MedGemma 27B reaches 0.850, Llama 4 Scout reaches 0.820, and Mistral Small 3.2 reaches 0.810, compared with 0.780 for GPT-5.6 Luna. Between-model ranking on this metric remains descriptive because the Wilson intervals overlap (see below). When examining the results across various fields, Llama 4 Scout and Mistral Small 3.2 often match GPT-5.6 Luna in accuracy. Genetics, lactation, and pregnancy show high accuracy across models, but this largely reflects the N/A$=$Suitable comparison policy and the prevalence of non-risk labels; macro-F1 on genetics is near 0.49 for three of the four models. MedGemma 27B reaches 1.000 accuracy on genetics and lactation without RAG, and GPT-5.6 Luna, Llama 4 Scout, and Mistral Small 3.2 likewise reach 1.000 on lactation. Pregnancy accuracy is 0.920--0.930 without RAG. Additionally, Llama 4 Scout's comorbidity accuracy (0.910) is close to that of GPT-5.6 Luna (0.940).

Grounding retrieved SmPC text can help on some safety-critical fields, but it is not a uniform improvement. Mean field accuracy rises modestly with RAG for all four models: GPT-5.6 Luna from 0.820 to 0.824, Llama 4 Scout from 0.785 to 0.804, MedGemma 27B from 0.754 to 0.759, and Mistral Small 3.2 from 0.808 to 0.811. The largest field-level gains are in contraindications (Llama 4 Scout 0.550 to 0.750; Mistral Small 3.2 0.690 to 0.940) and dose (all four models, for example GPT-5.6 Luna 0.640 to 0.820). Comorbidity and warning accuracy decline for several models. Overall-suitability accuracy is higher with RAG for every model: 0.780 and 0.930 for GPT-5.6 Luna, 0.820 and 0.900 for Llama 4 Scout, 0.850 and 0.930 for MedGemma 27B, and 0.810 and 0.890 for Mistral Small 3.2, without and with RAG, respectively. Retrieved passages, therefore, help when used for dose and interaction constraints, but they can also be misapplied.

\paragraph{Uncertainty and factual grounding}
Binomial 95\% Wilson intervals for overall-suitability accuracy ($n=100$ profiles) are GPT-5.6 Luna $0.780$ $[0.689, 0.850]$ without RAG and $0.930$ $[0.863, 0.966]$ with RAG; Llama 4 Scout $0.820$ $[0.733, 0.883]$ / $0.900$ $[0.826, 0.945]$; MedGemma 27B $0.850$ $[0.767, 0.907]$ / $0.930$ $[0.863, 0.966]$; Mistral Small 3.2 $0.810$ $[0.722, 0.875]$ / $0.890$ $[0.814, 0.937]$. The GPT-5.6 Luna RAG versus no-RAG intervals do not overlap; the other three models' paired intervals still overlap, and intervals overlap across models within each condition, therefore, between-model ranking on this metric remains descriptive. In the non-RAG setting, errors concentrate on warnings and, for some models, contraindications and dose, consistent with incomplete parametric recall of SmPC constraints. RAG reduces some of these gaps (notably dose and contraindications) but also increases N/A predictions on fields where the reference labels never use N/A, indicating that retrieved passages can be misapplied rather than simply omitted. We do not report a separate hallucination rate, because the evaluation is a constrained three-way classification against reference labels rather than an open-ended factuality audit.

\begin{figure*}
    \centering
    \includegraphics[width=\linewidth]{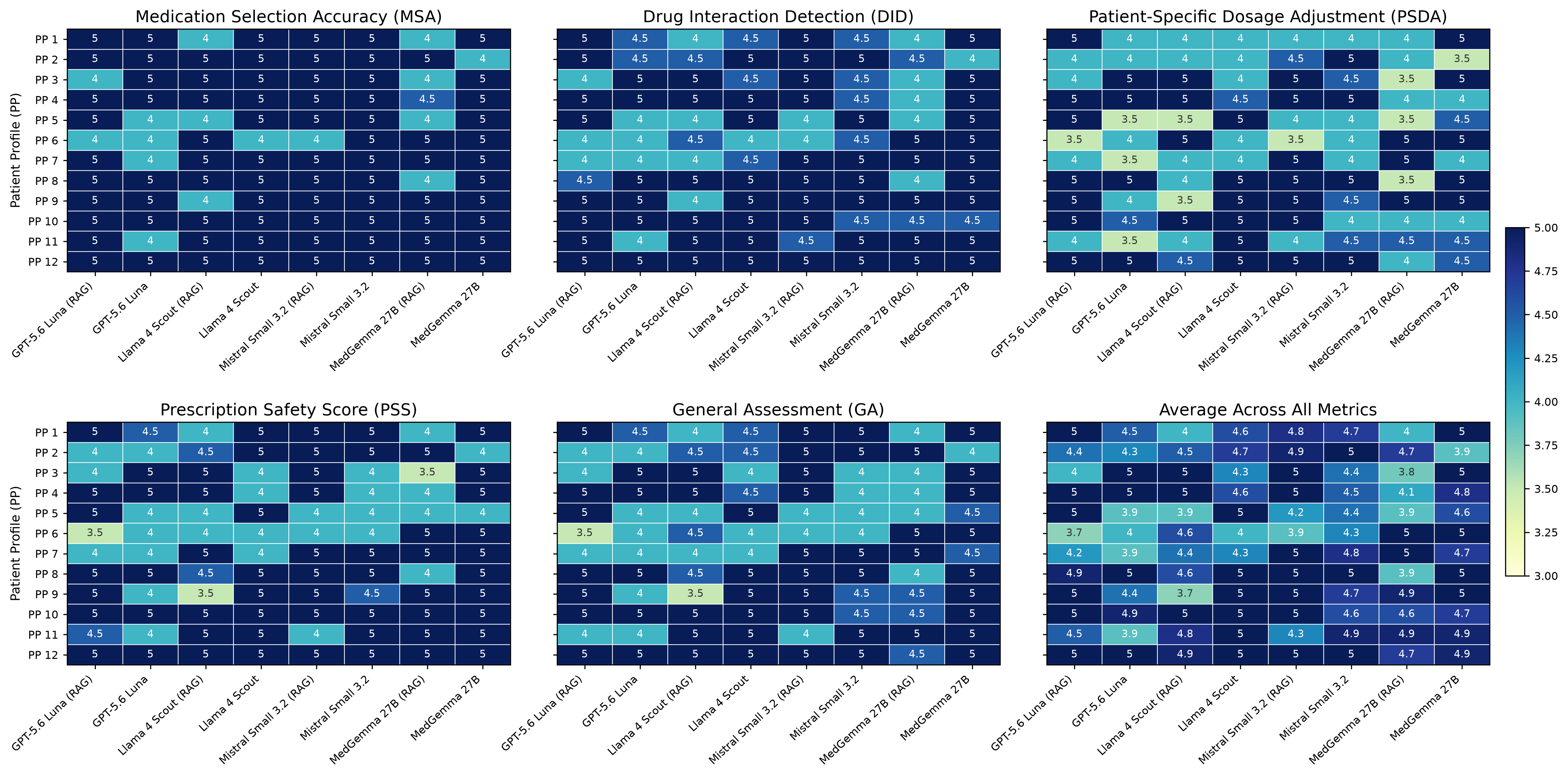}
    \caption{Two-reviewer mean scores (scale 0--5) for each of the 12 reviewed profiles. Each subplot is one rating dimension; the final panel averages the five dimensions.}
    \label{fig:qual_results}
\end{figure*}

\paragraph{Run-to-run stability}
As a secondary sensitivity analysis, we compared the two available repetitions for GPT-5.6 Luna, Llama 4 Scout, MedGemma, and Mistral Small 3.2. Mean field-level agreement between repetitions was 0.926--0.950 for GPT-5.6 Luna, 0.944--0.950 for Llama 4 Scout, 0.910 for MedGemma, and 0.971--0.981 for Mistral, across the RAG and no-RAG conditions. The principal RAG patterns were reproduced: mean field accuracy with RAG was at least as high as without RAG in both repetitions, and overall-suitability accuracy was higher with RAG in both repetitions for all four models (Llama 4 Scout $+8$ and $+7$ percentage points). Repetitions were not pooled as independent observations per patient.

\subsubsection{Qualitative Results}

To evaluate the clinical significance of personalized prescriptions generated by LLMs, we used two reviewers with different levels of training and clinical experience: a pharmacy student and a clinician co-author who remained actively involved throughout the project and had more than 10 years of experience managing multimorbidity and polypharmacy. Both reviewers were qualified to apply the study rubric to medication suitability, drug interactions, dose adjustments, and prescription safety. The assessments were conducted independently of one another and without access to the other reviewer's ratings. The workflow was designed to simulate a clinical decision-support setting rather than autonomous prescribing. Human review was maintained throughout the assessment process, consistent with the study's research-ethics requirements and with the MHRA's emphasis on human factors and safe human interaction with software-based medical devices \cite{MHRA_AIMD_2024}.Below, we describe the blinded two-rater assessment protocol and present the resulting findings.

\paragraph{Evaluation Strategy for Subjective Assessment}
For the subjective evaluation, we deterministically selected 12 of the 100 patient profiles, stratified across all five medications and, where available, both expert-labeled overall-suitability classes. Distinct risk scenarios were used to fill the medication-specific quotas. For each profile, we included the primary response from all four models under both RAG conditions, yielding 96 responses. Response order was randomized, and model and condition identities were concealed. Both reviewers independently evaluate each response using the following dimensions:

\begin{itemize}
    \item \textit{Medication Selection Accuracy (MSA)}: How appropriate is the model's judgment of the proposed medication for this patient, compared with expert assessment?
    \item \textit{Drug Interaction Detection (DID)}: How effectively does the model identify potential drug interactions for this patient, compared with the expert's judgment?
    \item \textit{Patient-Specific Dosage Adjustment (PSDA)}: How accurately does the model flag dosing concerns for special populations such as elderly patients or those with renal impairment?
    \item \textit{Prescription Safety Score (PSS)}: How safe is the model's overall suitability assessment, judged as input to a prescribing decision?
\end{itemize}

In addition to evaluating the aforementioned critical features, both reviewers are asked to provide a general assessment of the LLMs' overall effectiveness in generating safe, patient-centered prescriptions, without focusing on any specific feature (e.g., MSA, DID, PSDA, or PSS). To quantify the subjective assessment, we use a six-level scale from 0 to 5, where 0 indicates inappropriate and 5 indicates highly appropriate.

\begin{table*}
\centering
\scriptsize
\caption{Mean $\pm$ SD of the two-reviewer mean score across 12 patient profiles (scale 0--5; observed scores 3--5). RAG = with Retrieval-Augmented Generation; No RAG = without.}
\label{tab:qual_summary}
\scalebox{0.99}{
\begin{tabular}{llccccc}
\toprule
\textbf{Model} & \textbf{Cond.} & \textbf{MSA} & \textbf{DID} & \textbf{PSDA} & \textbf{PSS} & \textbf{GA} \\
\midrule
 GPT-5.6 Luna & RAG    & $4.83 \pm 0.39$ & $4.71 \pm 0.45$ & $4.54 \pm 0.58$ & $4.58 \pm 0.56$ & $4.54 \pm 0.58$ \\
& No RAG & $4.67 \pm 0.49$ & $4.58 \pm 0.47$ & $4.25 \pm 0.62$ & $4.46 \pm 0.50$ & $4.46 \pm 0.50$ \\
\midrule
Llama 4 Scout & RAG    & $4.75 \pm 0.45$ & $4.58 \pm 0.47$ & $4.29 \pm 0.58$ & $4.54 \pm 0.54$ & $4.50 \pm 0.52$ \\
& No RAG & $4.92 \pm 0.29$ & $4.79 \pm 0.33$ & $4.54 \pm 0.50$ & $4.67 \pm 0.49$ & $4.62 \pm 0.43$ \\
\midrule
Mistral Small 3.2 24B Instruct & RAG    & $4.92 \pm 0.29$ & $4.79 \pm 0.40$ & $4.58 \pm 0.56$ & $4.75 \pm 0.45$ & $4.75 \pm 0.45$ \\
& No RAG & $5.00 \pm 0.00$ & $4.79 \pm 0.26$ & $4.46 \pm 0.45$ & $4.62 \pm 0.48$ & $4.58 \pm 0.47$ \\
\midrule
MedGemma 27B Text IT & RAG    & $4.62 \pm 0.48$ & $4.50 \pm 0.48$ & $4.17 \pm 0.58$ & $4.54 \pm 0.58$ & $4.46 \pm 0.45$ \\
& No RAG & $4.92 \pm 0.29$ & $4.88 \pm 0.31$ & $4.50 \pm 0.52$ & $4.83 \pm 0.39$ & $4.83 \pm 0.33$ \\
\bottomrule
\end{tabular}}
\end{table*} 

\paragraph{Results using Subjective Assessment}

Table~\ref{tab:qual_summary} reports the mean and standard deviation of the two-reviewer mean score across the 12 evaluated profiles for each model and condition. Figure~\ref{fig:qual_results} visualizes the per-profile scores. Across the 96 response items rated by both reviewers, quadratic-weighted Cohen's kappa was 0.96 for MSA, 0.59 for DID, 0.69 for PSDA, 0.84 for PSS, and 0.69 for GA. Exact agreement was 99\%, 82\%, 77\%, 92\%, and 83\%, respectively. No paired scores differed by more than one point. These results indicate very high agreement for MSA and PSS and moderate-to-substantial agreement for DID, PSDA, and GA. The lower kappa for DID despite relatively high exact agreement may reflect the concentration of ratings in the upper categories: 98\% of all DID ratings were 4 or 5. The reported statistics are based on the independent ratings of the pharmacy student and experienced clinician before any discussion or reconciliation. Accordingly, the qualitative results should be interpreted as structured reviewer assessments rather than as a substitute for evaluation by a larger clinical panel.

Reviewers scored most responses in the upper part of the scale. Under that ceiling, RAG did not raise every model's mean on every dimension in Table~\ref{tab:qual_summary}. GPT-5.6 Luna was the only model with a higher two-reviewer mean on all five dimensions with RAG than without it, with the largest gain on PSDA (4.25 to 4.54). Mistral Small 3.2 showed small RAG gains on PSDA, PSS, and GA, while MSA was already at ceiling without RAG ($5.00$). Llama 4 Scout and MedGemma 27B received higher means without RAG across all dimensions. MedGemma with RAG had the lowest overall qualitative mean (4.46), including the lowest PSDA (4.17). On the 12 reviewed profiles, therefore, the reviewer scores do not track mean field accuracy: Table~\ref{tab:results} shows modest RAG gains in mean field accuracy, driven by dose and contraindications, whereas expert ratings did not rise uniformly with RAG. DID ratings should not be treated as a proxy for the contraindication field.

Among RAG conditions, Mistral Small 3.2 had the highest mean across the five dimensions (4.76), ahead of GPT-5.6 Luna (4.64) and Llama 4 Scout (4.53). Averaging those dimensions, MedGemma without RAG had the highest mean in the 12-profile sample (4.79). Standard deviations (0.00--0.62) are large relative to most between-condition gaps and we treat the ranking as descriptive. This qualitative review used 12 of the 100 synthetic profiles and five medications, limiting statistical generalization. All data were English-based. Quadratic-weighted kappa and exact agreement indicate that the two reviewers ranked responses similarly, but the small review panel remains a limitation, warranting larger, multilingual, multi-reviewer studies in future work.

\section{Conclusions}
\label{sec:concs}

In this paper, we have made three major contributions. First, we survey current developments and challenges in open Large Language Models (LLMs) and Artificial Intelligence Foundation Models (AIFMs). Second, we develop a taxonomy that organizes these models by application area and by degree of release openness. Third, we present a case study focused on LLM-empowered personalized prescriptions, in which we analyze the feasibility of using open-weight LLMs for critical clinical tasks such as prescribing and compare their performance with a proprietary baseline. Our exploration of open LLMs and AIFMs in healthcare highlights a pivotal shift towards more accessible and transparent AI technologies. Our comprehensive survey and taxonomy provide a structured overview of the current landscape, identifying both the strengths and limitations of open LLMs and AIFMs within healthcare applications. 

Two tensions emerge from our findings, which we frame as paradoxes. The first is a \emph{grounding paradox}. Across 100 controlled synthetic profiles, the evaluated open-weight models achieved overall suitability accuracy comparable to, and in some cases exceeding, the proprietary baseline, and RAG improved overall accuracy for every model. RAG also improved dose and contraindication accuracy, but comorbidity and warning accuracy often declined. In the 12-profile blinded review by two raters (a pharmacy student and a clinician co-author; quadratic-weighted kappa 0.59--0.96 across dimensions), RAG lowered the ratings of two of the four models (Llama 4 Scout and MedGemma 27B Text IT) on all five dimensions. Label-match metrics and expert ratings should therefore be read as complementary rather than interchangeable, and these findings motivate evaluating retrieval quality and evidence use, rather than assuming that grounding alone improves clinical quality.

The second is an \emph{openness paradox}. The models that appear most suitable for adaptation in healthcare often carry licenses that restrict the redistribution of adapted versions, while many models described as ``open'' in the literature do not meet the Open Source Initiative (OSI) definition of open source. Thus, the availability of model weights does not necessarily translate into openness in terms of modification, redistribution, or downstream use. Despite these limitations, open-weight LLMs and AIFMs remain promising for healthcare because they can offer greater opportunities for adaptation, customization, and local deployment than many proprietary alternatives. At the same time, these benefits introduce risks related to misuse, accountability, and governance. Their broader use in healthcare therefore requires clearer documentation of model and license conditions, careful assessment of downstream risks, and governance mechanisms that reflect the specific clinical context in which these models are deployed.

\section*{Acknowledgment}
The authors would like to acknowledge support from the Qatar University High Impact Internal Grant (QUHI-CENG23/24-127). The statements made herein are solely the responsibility of the authors.

\bibliographystyle{IEEEtran}

\end{document}